\documentclass[journal]{IEEEtran}
\usepackage[cmex10]{amsmath}
\usepackage{amssymb}
\usepackage{mdwmath}
\usepackage[T1]{fontenc}
\usepackage[noadjust]{cite}
\usepackage{graphicx}
\usepackage{mdwtab}
\usepackage{eqparbox}
\usepackage{multirow}
\usepackage{algorithm,algpseudocode}
\usepackage{diagbox}
\usepackage{threeparttable}
\usepackage{bm}
\usepackage{enumerate}
\usepackage{epstopdf}
\usepackage{subcaption}
\usepackage[colorlinks,linkcolor=red,anchorcolor=magenta,citecolor=red]{hyperref}
\usepackage{url}
\usepackage{makecell}
\usepackage{setspace}
\usepackage{xcolor}

\makeatletter

\newcommand{\Rmnum}[1]{\expandafter\@slowromancap\romannumeral #1@}
\makeatother
\usepackage{ntheorem}
\newtheorem{definition}{\textbf{Definition}}
\newtheorem{theorem}{\textbf{Theorem}}
\newtheorem{lemma}{\textbf{Lemma}}
\newtheorem{remark}{\textbf{Remark}}
\newtheorem{assumption}{\textbf{Assumption}}
\begin{document}
	\title{Understanding Short-Range Memory Effects in Deep Neural Networks}
	\author{Chengli Tan, Jiangshe Zhang, and Junmin Liu
		\thanks{This work was supported by the National Key Research and Development Program of China under Grant 2020AAA0105601 and by the National Natural Science Foundation of China under Grants 61976174, 61877049.}
		\thanks{The authors are with the School of Mathematics and Statistics, Xi'an Jiaotong University, Xi'an 710049, China.}}
	\markboth{IEEE Transactions on Neural Networks and Learning Systems}{C. Tan \emph{et al.}: Understanding Short-Range Memory Effects in Deep Neural Networks}
	\maketitle
	\begin{abstract}
		Stochastic gradient descent (SGD) is of fundamental importance in deep learning. Despite its simplicity, elucidating its efficacy remains challenging. Conventionally, the success of SGD is ascribed to the stochastic gradient noise (SGN) incurred in the training process. Based on this consensus, SGD is frequently treated and analyzed as the Euler-Maruyama discretization of stochastic differential equations (SDEs) driven by either Brownian or L\'evy stable motion. In this study, we argue that SGN is neither Gaussian nor L\'evy stable. Instead, inspired by the short-range correlation emerging in the SGN series, we propose that SGD can be viewed as a discretization of an SDE driven by fractional Brownian motion (FBM). Accordingly, the different convergence behavior of SGD dynamics is well-grounded. Moreover, the first passage time of an SDE driven by FBM is approximately derived. The result suggests a lower escaping rate for a larger Hurst parameter, and thus SGD stays longer in flat minima. This happens to coincide with the well-known phenomenon that SGD favors flat minima that generalize well. Extensive experiments are conducted to validate our conjecture, and it is demonstrated that short-range memory effects persist across various model architectures, datasets, and training strategies. Our study opens up a new perspective and may contribute to a better understanding of SGD. 
	\end{abstract}
	\section{Introduction}
	\IEEEPARstart{D}{espite} the great uncertainty of stochastic processes that widely occur in nature, it is recognized that they have memories as well. In the early 1950s, considering regular floods and irregular flows, which were a severe impediment to development, the Egyptian government decided to construct a series of dams and reservoirs to control the Nile. To devise a method of water control, several hydrologists examined the rain and drought volatility of the Nile and discovered a short-range relationship therein \cite{hurst1951long, black1965long}. This implies that, contrary to the common independence assumption, there is non-negligible dependence between the present and the past. Determining the behavior and predictability of stochastic systems is critical to understanding, for example, climate change \cite{yuan2014extracting}, financial market \cite{cont2005long}, and network traffic \cite{grossglauser1999relevance, karagiannis2004long}. Accordingly, it is natural to ask whether or not similar memory effects exist in deep neural networks (DNNs), which can be seen as complex stochastic systems.
	\par
	Generally, many tasks in DNNs can be reduced to solving the following unconstrained optimization problem
	\begin{equation}
		\min_{w\in\mathbf{R}^d} f(w) = \frac{1}{N} \sum_{i=1}^{N}  f_{i}(w),\label{eq:obj}
	\end{equation}
	where $w\in\mathbf{R}^d$ represents the weights of the neural network, $f$  represents the loss function, which is typically non-convex in $w$, and each $i \in \{1, 2, \cdots, N\}$ corresponds to an instance tuple $(x_i, y_i)$ of the training samples. Due to the limited resources and their remarkable inefficiency, deterministic optimization algorithms, such as gradient descent (GD), cannot easily train models with millions of parameters. Instead, in practice, stochastic optimization approaches, such as SGD and its variants AdaGrad \cite{duchi11a} and Adam \cite{kingma2014adam}, are employed. 
	The SGD recursion process is defined as 
	\begin{equation}
		w_{k+1} = w_{k} - \eta \nabla\widetilde{f}(w_k),\label{eq:sgd_update}
	\end{equation}
	where $k$ denotes the $k$th step, $\eta$ is the learning rate, and $\nabla\widetilde{f}(w_k)$ is the estimation of the true gradient at the current step, which is given by
	\begin{equation}
		\displaystyle
		\nabla\widetilde{f}(w_k) = \frac{1}{n}\sum_{i \in \mathcal{S}_k}\nabla f_i(w_k).
	\end{equation}
	Here $\mathcal{S}_k$ is an index set of a batch of samples independently drawn from the training dataset $\mathcal{S}$, and $n=|\mathcal{S}_k|$ denotes the cardinality of $\mathcal{S}_k$.
	\par
	Despite being overwhelmingly over-parameterized, SGD prevents DNNs from converging to local minima that cannot generalize well \cite{keskar2016large, zhang2016understanding, hoffer2017train, jastrzkebski2017three, zhu2018anisotropic}. One point of view is that SGD induces implicit regularization by incurring SGN during the training process \cite{bottou2018optimization}, which is defined as 
	\begin{equation}
		\nu_k = \nabla\widetilde{f}(w_k) - \nabla{f}(w_k),
	\end{equation}
	where $\nabla{f}(w_k)$ represents the gradient of  the loss function over the full training dataset $\mathcal{S}$
	\begin{equation}
		\displaystyle
		\nabla{f}(w_k) = \frac{1}{N}\sum_{i=1}^N\nabla f_i(w_k).
	\end{equation}
	This claim is supported by several observations found in numerous experiments:
	\begin{enumerate}
		\item SGD generally outperforms GD \cite{zhu2018anisotropic};
		\item SGD using large batch size often leads to performance deterioration compared with SGD using small batch size \cite{jastrzkebski2017three, smith2017don, yao2018hessian, he2019control};
		\item SGD approximations are not as effective as SGD, even if they are tuned to the same noise intensity \cite{zhu2018anisotropic}.
	\end{enumerate}
	\par
	The first observation confirms that the existence of SGN is a requisite for models to avoid getting trapped in suboptimal points. The second observation indicates that the model performance is heavily influenced by the noise intensity. The last observation further implies that the model performance is affected not only by the noise intensity, but also by the noise shape. It should be clarified that SGN arises in the process of batch sampling, which is the major source of randomness, except for weight initialization and dropout operations. Let
	\begin{equation}
		\nabla\widetilde{f}(w_k) = \frac{1}{N}\nabla{f}_k\cdot \varrho, 
	\end{equation}
	where $\nabla{f}_k=(\nabla f_1(w_k), \cdots, \nabla f_N(w_k))\in \mathbf{R}^{d\times N}$, $\varrho \in \mathbf{R}^N$ is a random vector, indicating which samples are selected at the $k$th iteration. Let $1$ be the matrix the elements of which are equal to 1; then, we can define the sampling noise as $\zeta = \frac{1}{N}(\varrho - 1)$. The properties of  $\zeta$ are given as follows \cite{wu2019noisy}:
	\begin{enumerate}
		\item For a batch sampled without replacement, the SGD sampling noise $\zeta$ satisfies
		\begin{equation}
			\mathbb{E}\left[\zeta\right]=0,~\operatorname{Var}\left[\zeta\right]=\frac{N-n}{n N(N-1)}\left(\mathrm{I}-\frac{1}{N} 1 1^{T}\right),\label{eq:sampling noise a}
		\end{equation}
		\item For a batch sampled with replacement, the SGD sampling noise $\zeta$ satisfies
		\begin{equation}
			\mathbb{E}\left[\zeta\right]=0,~\operatorname{Var}\left[\zeta\right]=\frac{1}{n N}\left(\mathrm{I}-\frac{1}{N} 1 1^{T}\right).
			\label{eq:sampling noise b}
		\end{equation}
	\end{enumerate}
	Here $\mathrm{I}\in \mathbf{R}^{N\times N}$ is an identity matrix. 
	\par
	From the perspective of sampling noise, the SGN $\nu_k$ is equivalent to the product of the gradient matrix $\nabla{f}(w_k)$ and the corresponding sampling noise $\zeta$, \textit{i.e.},
	\begin{equation}
		\nu_k = \nabla\widetilde{f}(w_k) - \nabla{f}(w_k) = \nabla{f}_k\cdot \zeta.
	\end{equation}
	Therefore, the first two moments of $\nu_k$ are given by
	\begin{equation}
		\mathbb{E}\left[\nu_k\right] =  \nabla{f}_k\cdot\mathbb{E}\left[ \zeta \right] = 0,
	\end{equation}
	and 
	\begin{equation}
		\mathrm{Var}\left[\nu_k\right] = \nabla{f}_k \mathrm{Var}\left[\zeta\right]\nabla{f}_k^T\triangleq\Sigma_k.
		\label{eq:sgd-covariance}
	\end{equation}
	It is noteworthy that although the sampling noise is time-independent, the SGN is deeply affected by the present parameter state.
	\par
	A popular approach to studying the behavior of SGN is to view SGD as a discretization of SDEs \cite{jastrzkebski2017three, mandt2017stochastic, li2017stochastic, chaudhari2018stochastic, hu2019diffusion}. This approach relies on the assumption that if the batch size $|\mathcal{S}_k|$ is sufficiently large, then by invoking the central limiting theorem (CLT), the SGN follows a Gaussian distribution
	\begin{equation}
		\nu_k \sim \mathcal{N}(0, \Sigma_k).
	\end{equation}
	Based on this assumption, (\ref{eq:sgd_update}) can be translated into
	\begin{equation}
		w_{k+1}=w_{k}-\eta \nabla f\left(w_{k}\right)+\eta \nu_k.
		\label{eq:discrete-sgd}
	\end{equation}
	If the learning rate $\eta$ is sufficiently small to be treated as a time increment, (\ref{eq:discrete-sgd}) implements an Euler-Maruyama approximation of the following It\^{o} process \cite{jastrzkebski2017three}
	\begin{equation}
		\mathrm{d} w_{t}=-\nabla f\left(w_{t}\right) \mathrm{d} t+\sqrt{\eta}R_k \mathrm{d} \mathrm{B}_{t},
		\label{eq:maruyama}
	\end{equation}
	where $R_kR_k^T=\Sigma_k$ and $\mathrm{B}_{t}$ is the standard Brownian motion.
	Following this line, He et al. \cite{he2019control} argued that the generalization gap is governed by the ratio of the learning rate to the batch size, which directly controls the intensity of SGN. Further, they related this ratio to the flatness of minima obtained by SGD \cite{keskar2016large, hinton1993keeping, hochreiter1997flat, chaudhari2019entropy}, and concluded that a larger ratio implies that the minima are wider and thus generalize better. Another striking perspective is that SGD simulates a Markovian chain with a stationary distribution \cite{mandt2017stochastic}. Under certain condition that the loss function can be well approximated by a quadratic function, SGD can be used as an approximate Bayesian posterior inference algorithm. It follows that SGD can be viewed as a discretization of the continuous-time Ornstein--Uhlenbeck (OU) process \cite{mandt2017stochastic}
	\begin{equation}
		\mathrm{d}w_t = -\eta A w_t\mathrm{d}t + \frac{\eta}{\sqrt{n}}B \mathrm{d}B_t,
		\label{eq:ou-sde}
	\end{equation}
	which has the following analytic stationary distribution
	\begin{equation}
		q(w) \propto \exp \left\{-\frac{1}{2} w^{T} \Sigma^{-1} w\right\},
	\end{equation}
	where $A, B, \Sigma$ satisfy
	\begin{equation}
		\Sigma A + A \Sigma = \frac{\eta}{n}BB^T.
	\end{equation}
	\begin{figure*}[t]
		\centering
		\begin{subfigure}[b]{0.48\textwidth}
			\centering
			\includegraphics[width=0.8\textwidth, clip, trim= 0 0 0 0]{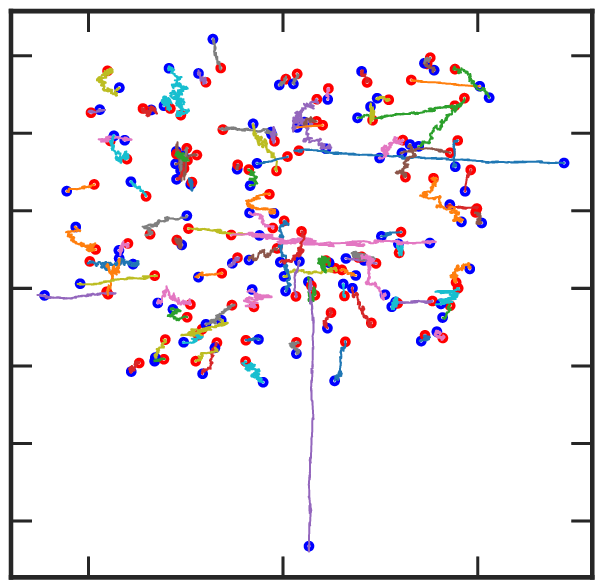}
			\caption{Batch Size = 64}
		\end{subfigure}
		\begin{subfigure}[b]{0.48\textwidth}
			\centering
			\includegraphics[width=0.8\textwidth, clip, trim= 0 0 0 0]{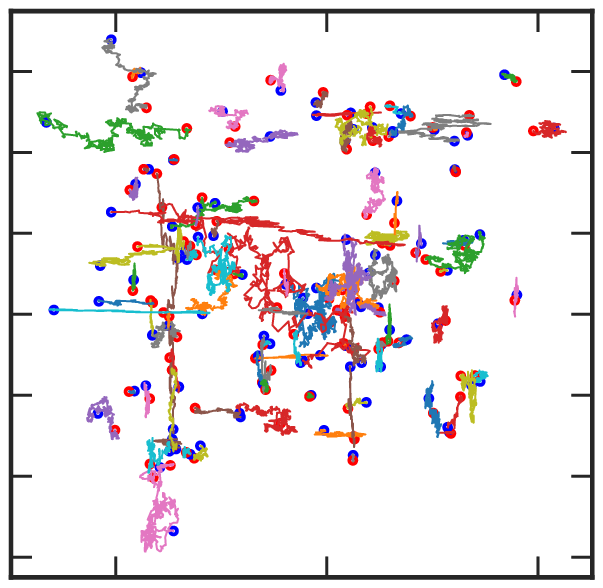}
			\caption{Batch Size = 512}
		\end{subfigure}
		\caption{Local trajectories of SGN mapped onto two-dimensional space for different batch sizes. In both cases, the noise is generated from the last iteration, and two adjacent dimensions are bound together to plot one trajectory line. The starting point is uniformly sampled in $[0, 1]$. Trajectories that resemble lines indicate that the noise in one of the two dimensions is near zero. The red empty circle indicates the starting point of the trajectory, whereas the blue circle indicates the ending point of the trajectory.}
		\label{fig:trajectory of noise}
	\end{figure*}
	Furthermore, it was recently argued that the critical assumption that SGN follows a Gaussian distribution does not necessarily hold  \cite{simsekli2019tail, nguyen2019first}. Instead, it was assumed that the second moment of SGN may not be finite, and it was concluded that the noise follows a L\'evy random process by invoking the generalized CLT. Therefore, it was proposed that SGD can be modeled as a L\'evy-driven SDE \cite{simsekli2019tail}
	\begin{equation}
		\mathrm{d} w_{t}=-\nabla f\left(w_{t}\right) \mathrm{d} t+\eta^{(\alpha-1) / \alpha} \sigma(t, w_t) \mathrm{d} \mathrm{L}_{t}^{\alpha},
		\label{eq:levy-sgd}
	\end{equation}
	where $\mathrm{L}_{t}^{\alpha}$ denotes the $d$-dimensional $\alpha$-stable L\'evy motion with independent components.
	When $\alpha=2$, (\ref{eq:levy-sgd}) reduces to a standard SDE driven by Brownian motion. It was argued that when $\alpha<2$, L\'evy-driven GD dynamics require only polynomial time to transit to another minimum (rather than exponential time, as in Brownian motion-driven GD dynamics) because the process can incur discontinuities, and thus the algorithm can directly jump out.
	\par
	Instead of focusing on the influence of noise intensity and noise class, Zhu et al. \cite{zhu2018anisotropic} studied noise shape, namely, the covariance structure. According to the classical statistical theory \cite{pawitan2001all}, there is an exact equivalence between the Hessian matrix of the loss and the Fisher information matrix, namely, the covariance at the optimum. Practically, when the present parameter is around the true value, the Fisher matrix is close to the Hessian. Owing to the dominance of noise in the later stages of SGD iteration \cite{shwartz2017opening}, the gradient covariance is approximately equal to the Hessian. Although this is disputable \cite{dinh2017sharp}, it is widely appreciated that SGD favors flat minima, where the spectrum of the Hessian matrix is relatively small. By measuring the alignment of the noise covariance and the Hessian matrix of the loss function, it was concluded that anisotropic noise is superior to isotropic noise in terms of jumping out of sharp minima. More recently, based on the assumption that SGN follows a Gaussian distribution, Xie et al. \cite{xie2021diffusion} further derived the mean escaping time of Kramers problem and found that SGD favors flat minima exponentially more than sharp minima.
	\subsection{Issues}
	Although the aforementioned studies have confirmed the vitality of SGN, a satisfactory understanding of the full SGD dynamics is lacking.
	Several relevant issues are explained in detail below.
	\par
	The first issue relates to the infiniteness of the second moment of SGN, as assumed in \cite{simsekli2019tail, nguyen2019first}. According to \eqref{eq:sgd-covariance}, the noise covariance $\Sigma_k$ is the product of the sampling noise and the gradient matrix.
	Given the precondition that the covariance of the sampling noise $\zeta$ is finite, \eqref{eq:sgd-covariance} implies that whether or not the noise covariance $\Sigma_k$ is finite depends solely on the gradient matrix $\nabla{f}_k$.
	The main components of DNNs, such as matrix and convolution operations, are twice differentiable almost everywhere. In addition, common techniques in deep learning, such as batch normalization \cite{ioffe2015batch}, near-zero initialization \cite{he2015delving}, and gradient clipping \cite{glorot2010understanding}, ensure that the parameters will not excessively drift from their initialization. Hence, it is reasonable to assume that the gradient matrix lies in a bounded region of interest, and the finiteness of the noise covariance is ensured.
	Thus, the assumption that SGN follows a L\'evy stable distribution might be flawed.
	\par
	The second issue involves the normality of SGN. Although the second moment of SGN is finite, in practice, it remains unclear whether the batch size is sufficiently large to ensure that the normality assumption holds. To this end, Panigrahi et al. \cite{panigrahi2019non} directly conducted Shapiro and Anderson normality tests on SGN. The results suggest that SGN is highly non-Gaussian and follows a Gaussian distribution only in the early training stage for batch sizes of 256 and above.
	\subsection{Contributions}
	Traditionally, we assume that SGN is independently generated at different iterations, or nearly so. Fig. \ref{fig:trajectory of noise} shows the two-dimensional trajectories of SGN of the training process for different batch sizes. It can be observed that the patterns are disparate, and the trajectories in Fig. \ref{fig:trajectory of noise}(a) are almost smooth and regular, whereas the trajectories in Fig. \ref{fig:trajectory of noise}(b) are fractal-like and exhibit rapid mutations. This raises the question of whether there exists strong inter-dependence between the SGN series.
	\par
		One important class of stochastic processes to characterize the inter-dependence of economic data is known as FBM.
		Bearing the observation in mind that the self-similar sample paths of FBM are fractals, in this study, we leverage FBM to study the inter-dependence of the SGN series. 
	Instead of assuming that SGN is Gaussian or follows a L\'evy random process, we hypothesize that it is generated by FBM, the increments of which are fractional Gaussian noise (FGN) and time-dependent.
	Using the same rationale as in (\ref{eq:maruyama}) and (\ref{eq:levy-sgd}), we reformulate the SGD dynamics as a discretization of an FBM-driven SDE.
	According to the values of the Hurst parameter, FBM can be classified as super-diffusive ($0.5<H<1$), sub-diffusive ($0<H<0.5$), and normal-diffusive ($H=0.5$).
	\begin{figure*}[t]
		\centering
		\includegraphics[width=0.8\textwidth, clip, trim= 0 0 0 0]{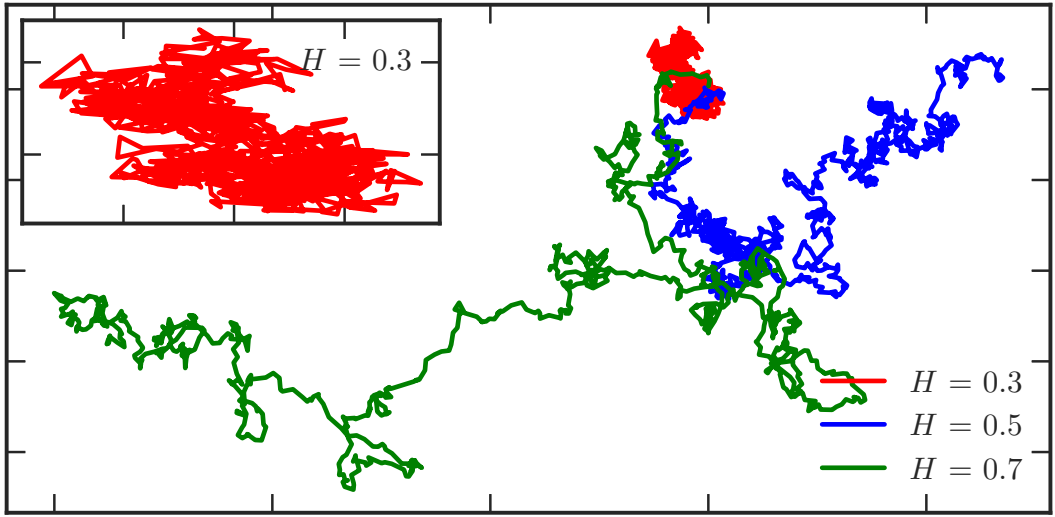}
		\caption{Two dimensional sample paths of FBM for $H=0.3$ (red), $H=0.5$ (blue) and $H=0.7$ (green). The inset shows a zoomed-in view of the trajectory for $H=0.3$.}
		\label{fig:sample path of fbm}
	\end{figure*}
	Therefore, an FBM-driven SDE exhibits a fundamentally different behavior from that of an ordinary SDE driven by Brownian motion.
	As described in Section \ref{sec:methodology}, the convergence rate of an FBM-driven SDE is closely related to the Hurst parameter; this accounts for the different convergence behaviors in the SGD training process. When the training process gets trapped in local minima, we further study the escaping efficiency from the perspective of the first passage time of an FBM-driven SDE.
	\par
	In general, the contributions of this study are the following:
	\begin{enumerate}
		\item In contrast with previous studies, we assume that the SGN is dependent on history, and based on this assumption, we propose to analyze SGD as a discretization of an FBM-driven SDE;
		\item We analytically derive the state probability density of an FBM-driven SDE for the first time and obtain an approximate first passage time probability density.
	\end{enumerate}
	\par
	We conduct extensive experiments to examine short-range memory effects in DNNs, and several important observations are recovered:
	\begin{enumerate}
		\item In all experiments, there is no evidence that SGN is Gaussian or L\'evy stable even in model architectures such as Long Short-Term Memory (LSTM) \cite{hochreiter1997long} and Vision Transformer (ViT) \cite{dosovitskiy2021image}, which are specially designed to handle with sequential datasets. Instead, it is well described by FBM;
		\item Hyperparameters such as batch size (BS) and learning rate (LR) have a great influence on the Hurst parameter. Generally, a large batch size leads to a small Hurst parameter, whereas a large learning rate corresponds to a large Hurst parameter;
		\item The Hurst parameter and the curvature at local minima are bundled together. A large Hurst parameter implies small eigenvalues of the Hessian matrix and this is aligned with the observation that SGD favors flat minima;
		\item Though not theoretically guaranteed, the findings of Hurst parameter on vanilla SGD are still true when employing different training strategies such as momentum and weight decaying.
	\end{enumerate}
	\par
	The remainder of this paper is organized as follows. In Section \uppercase\expandafter{\romannumeral2} we give some preliminaries. The methodology including convergence analysis and first passage time analysis is presented in Section \uppercase\expandafter{\romannumeral3}. And Section \uppercase\expandafter{\romannumeral4} presents the experimental results. We conclude in Section \uppercase\expandafter{\romannumeral5}.
	\section{Preliminaries}
	FBM was originally studied in \cite{kolmogorov1940curves, yaglom1955correlation}. It was later popularized through  \cite{mandelbrot1968fractional} and found a wide range of applications in Internet traffic \cite{leland1994self}, financial market \cite{campbell1997econometrics}, and fractal geometry \cite{mandelbrot1967long}.
	Informally, it can be seen as an extension of Brownian motion. We now provide its formal definition and discuss several important properties of FBM.
	\begin{definition}
		Given a complete probability space $(\Omega, \mathcal{F}, P)$, FBM is an almost surely continuous centered Gaussian process $\{B_t^H, t\geq 0\}$ with covariance function
		\begin{equation}
			\mathbb{E}[B_t^H B_s^H] = \frac{1}{2} \left(t^{2H} + s^{2H} - (t-s)^{2H}\right),\label{eq:fbm}
		\end{equation}
		where $H$ is a real number in $(0, 1)$, called the Hurst index or Hurst parameter associated with the FBM.
	\end{definition}
	This process has only one parameter, \textit{i.e.}, the Hurst parameter $H$. It should be noted that when $H = 0.5$, FBM reduces to the standard Brownian motion. If $H \neq 0.5$, the process is neither Markovian nor semi-martingale, and therefore useful analytic techniques from stochastic calculus cannot be directly adapted to its analysis. The definition readily implies several important properties that separate FBM from Brownian motion.
	\par
	\emph{\textbf{Self-similarity}.} Given $a > 0$, the process $\{B_{at}^H, t\geq0\}$ has the same covariance and thus the same distribution as $\{a^HB_t^H, t\geq0\}$. This property is called self-similarity with parameter $H$, implying that the behavior of the random process remains the same in each time interval if scaled properly.
	\par
	\emph{\textbf{Stationary increments}.} For any fixed $s$, it follows from (\ref{eq:fbm}) that the covariance function of $\{B_{t+s}^H - B_s^H, t\geq0\}$ is the same as that of $\{B_t^H, t\geq0\}$. 
	\par
	\emph{\textbf{Dependence of increments}.} We assume that the time intervals $[s, t]$ and $[m, n]$ are disjoint. Without loss of generality, we assume that $s<t<m<n$. It can be directly inferred from (\ref{eq:fbm}) that
	\setlength{\arraycolsep}{0.0em}
	\begin{eqnarray}
		\mathbb{E}\left[\left(B_{t}^{H}-B_{s}^{H}\right)\left(B_{m}^{H}-B_{n}^{H}\right)\right] 
		&{}={}&\frac{1}{2}(\left|t-m\right|^{2 H}+\left|n-s\right|^{2 H}\nonumber\\
		&&{-}\:\left|n-t\right|^{2 H}-
		\left|m-s\right|^{2 H}),\nonumber
	\end{eqnarray}
	\setlength{\arraycolsep}{5pt}
	yielding
	\begin{equation}
		\mathbb{E}\left[\left(B_{t}^{H}-B_{s}^{H}\right)\left(B_{m}^{H}-B_{n}^{H}\right)\right] < 0, \quad\text{if\ } H \in (0, 0.5),
	\end{equation}
	and
	\begin{equation}
		\mathbb{E}\left[\left(B_{t}^{H}-B_{s}^{H}\right)\left(B_{m}^{H}-B_{n}^{H}\right)\right] > 0, \quad\text{if\ } H \in (0.5, 1).
	\end{equation}
	Therefore, for $H \in (0, 0.5)$, FBM exhibits short-range dependence, implying that if it was increasing in the past, it is more likely to decrease in the future and vice versa.
	By contrast, for $H \in (0.5, 1)$, FBM exhibits long-range dependence, that is, it tends to maintain the current trend.
	\par
	Fig. \ref{fig:sample path of fbm} shows several typical sample paths of FBM  on the plane for $H=0.3$, $H=0.5$, and $H=0.7$.
	It can be seen that when the Hurst parameter is small, the sample path exhibits an overwhelmingly large number of rapid changes. By contrast, the sample path appears significantly smoother when the Hurst parameter is large. It should also be noted that when the Hurst parameter is small, the random walker remains constrained in a narrow space.
	By contrast, the random walker explores a considerably broader space when the Hurst parameter is large.
	\section{Methodology}
	\label{sec:methodology}
	In contrast with (\ref{eq:maruyama}), we view SGD as a discretization of the following FBM-driven SDE
	\begin{equation}
		\mathrm{d} w_{t}=-\nabla f\left(w_{t}\right) \mathrm{d} t+\sigma(t, w_t) \mathrm{d}B_t^H,
		\label{eq:fbm-sgd}
	\end{equation}
	where $B_t^H$ stands for FBM with Hurst parameter $H$.
	For convenience, we use the notation $g(t, w_t) \triangleq \nabla f\left(w_{t}\right)$ in the sequel and (\ref{eq:fbm-sgd}) is rewritten as
	\begin{equation}
		\mathrm{d} w_{t}=-g(t, w_t) \mathrm{d} t+\sigma(t, w_t) \mathrm{d}B_t^H.
		\label{eq:s-fbm-sgd}
	\end{equation}
	Although the properties of SDEs driven by Brownian motion have been extensively studied, their fractional counterpart has not attracted significant attention. In \cite{hairer2005ergodicity, rascanu2002differential}, the authors argued that (\ref{eq:fbm-sgd}) in a simpler form admits a unique stationary solution, and every adapted solution converges to this stationary solution algebraically. 
	As SGD is viewed as a discretization of its continuous-time limit SDE, we are more concerned with the speed of convergence of the discrete approximation to its continuous solution.
	\subsection{Convergence Analysis}
		Divide the time interval $[0, T]$ into $K$ equal subintervals and denote
		$\eta=T/K, \tau_k=kT/K=k\eta, k=0, 1, \ldots, K$.
	We consider the discrete Euler--Maruyama approximation of the solution of (\ref{eq:s-fbm-sgd})
	\begin{equation}
		\widetilde{w}_{\tau_{k+1}}^\eta = \widetilde{w}_{\tau_{k}}^\eta - g(\tau_{k}, \widetilde{w}_{\tau_{k}}^\eta)\eta + \sigma(\tau_k, \widetilde{w}_{\tau_{k}}^\eta)\Delta B_{\tau_k}^H, \quad \widetilde{w}_{0}^\eta = w_0^\eta,
	\end{equation}
	and its corresponding continuous interpolations
	\begin{equation}
		w_{t}^\eta = \widetilde{w}_{\tau_{k}}^\eta - g(\tau_{k}, \widetilde{w}_{\tau_{k}}^\eta)(t-\tau_k) + \sigma(\tau_k, \widetilde{w}_{\tau_{k}}^\eta)(B_t^H - B_{\tau_k}^H),
	\end{equation}
	where $t \in [\tau_k, \tau_{k+1}]$. The above expression can also be reformulated in integral form
	\begin{equation}
		w_{t}^\eta = w_0 - \int_{0}^{t}g(t_\mu, w_{t_\mu}^\eta)\mathrm{d}\mu + \int_{0}^{t}\sigma(t_\mu, w_{t_\mu}^\eta)\mathrm{d}B_\mu^H,
	\end{equation}
	where $t_\mu=\tau_{k_\mu}$ and $k_\mu = \max\{k: \tau_k\leq \mu\}$.
	We now present several assumptions on the drift coefficient $g(t, w_t)$ and the diffusive coefficient $\sigma(t, w_t)$.
	\begin{definition}
		$f$ is said to be locally Lipschitz continuous if for every $N>0$, there exists $L_N>0$ such that
		\begin{equation*}
			|f(x) - f(y)| \leq L_N |x-y|,~\text{for all}~|x|, |y| \leq N.
		\end{equation*}
	\end{definition}
	\begin{definition}
		$f$ is said to be local $\alpha$-H\"older continuous if for every $N>0$, there exists $M_N>0$ such that
		\begin{equation*}
			|f(x) - f(y)| \leq M_N |x - y|^\alpha, \text{ for all }|x|, |y| \leq N.
		\end{equation*}
	\end{definition}
	\begin{assumption}
		Assume that $\sigma(t, w)$ is differentiable in $w$, and there exist constants
		$C$, $M>0$, $H<\beta<1$, and $1/H-1\leq \alpha \leq 1$ such that
		\begin{enumerate}
			\item $|\sigma(t, w)| \leq C (1+|w|)$ for all $t\in [0, T]$;
			\item $\sigma(t, w)$ is $M$-Lipschitz continuous in $w$ for all $t\in [0, T]$;
			\item $\nabla_w\sigma(t, w)$ is locally $\alpha$-H\"older continuous in $w$ for all $t\in [0, T]$;
			\item $|\sigma(t, w) - \sigma(s, w)| + |\nabla_w\sigma(t, w)-\nabla_w\sigma(s, w)|\leq M|t-s|^\beta$ for all $s \in [0, T]$.\\[-0.2cm]
		\end{enumerate}
	\end{assumption}
	\begin{assumption}
		The gradient $g(t, w)$ is locally Lipschitz continuous in $w$, and there exist constants $L>0$ and $2H-1<\gamma\leq1$ such that
		\begin{enumerate}
			\item $|g(t, w)| \leq L(1+|w|)$ for all $t\in [0, T]$;
			\item $g(t, w)$ is $\gamma$-H\"older continuous in $w$ for all $t\in [0, T]$.
		\end{enumerate}
	\end{assumption}
	Given that $\sigma(t, w)$ and $g(t, w)$ satisfy the above assumptions, we have the following theorem from \cite{mishura2008rate}.
	\begin{theorem}
		\label{theorem1}
		Let 
		\begin{equation}
			\Delta_{t, s}(w, w^\eta) = |w_t - w^\eta_t - w_s + w^\eta_s|,
		\end{equation}
		then the norm of the difference $|w_t - w_t^\eta|$ equipped in certain Besov space
		\begin{equation}
			U^\eta\triangleq \sup_{0\leq s \leq T} \left(|w_s - w^\eta_s| + \int_{0}^{t_s}|\Delta_{u, s}(w, w^\eta) |(s-u)^{-\alpha-1}\mathrm{d}u\right) 
		\end{equation}
		satisfies the following:
		\begin{enumerate}
			\item For any $\epsilon>0$ and any sufficiently small $\rho>0$, there exist $\eta_0>0$ and $\Omega_{\epsilon, \rho, \eta_0}\subset \Omega$ such that $P(\Omega_{\epsilon, \rho, \eta_0})>1-\epsilon$,
			and for any $w \in \Omega_{\epsilon, \rho, \eta_0}$ and $\eta<\eta_0$, we have
			$U^\eta \leq C(w)\eta^{2H-1-\rho}$,
			where $C(w)$ only depends on $\rho$;
			\item If, additionally, the coefficients $\sigma(t, w)$ and $g(t, w)$ are bounded, then for any $\rho\in(2H-1, 1)$ there exist $C(w) < \infty$ almost surely such that $U^\eta \leq C(w)\eta^{2H-1-\rho}$, where $C(w)$ does not depend on $\eta$.
		\end{enumerate}
	\end{theorem}
	It is well-known that the Euler-Maruyama scheme for SDEs driven by Brownian motion has a strong order of convergence $0.5$. By contrast, the above theorem implies that the convergence rate varies with the Hurst parameter, radically separating FBM from Brownian motion. Moreover, as $\sigma(t, w)$ and $g(t, w)$ are bounded in DNNs, as discussed previously, the conclusion $U^\eta \leq C(w)\eta^{2H-1-\rho}$ holds uniformly. This indicates that the error between the discretization and the true solution gained from SGD recursion is upper bounded and scales as $\mathcal{O}(\eta^{2H-1-\rho})$. Thus, it is straightforward to conclude that when the Hurst parameter is large, the corresponding SGD recursion has a fast convergence rate.
	\subsection{First Passage Time Analysis}
\begin{figure}[t]
		\centering
		\includegraphics[width=0.48\textwidth]{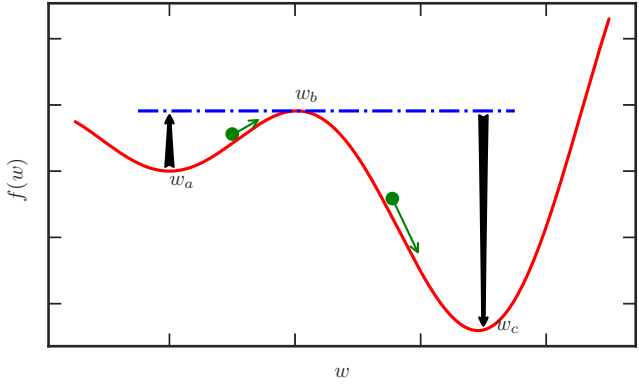}
		\caption{Potential function with two local minima $w_a$ and  $w_c$ separated by a local maximum $w_b$.}
		\label{fig:potential_function}
	\end{figure}

	In this section, we present our main contribution associated with the first passage time problem for FBM-driven SDEs. 
	The first passage time is the time $t$ at which a stochastic process passes over a barrier for the first time or escapes from an interval by reaching the absorbing boundaries. Suppose that $w_t$ is currently trapped in a suboptimal minimum $w_a$, as shown in Fig. \ref{fig:potential_function}, before sliding to another minimum $w_c$, we are interested in the time required to climb up to $w_b$ from the bottom, as in metastability theory \cite{hanggi1986escape}.
	This is a key problem in stochastic processes.
	However, except for Brownian motion and several particular random processes, there are few analytical results regrading to the Gaussian processes.
	\begin{assumption}
		Assume that the stationary distribution of the SGD recursion is constrained to a local region where the loss function is well approximated by a quadratic function
		\begin{equation}
			f(w) = \frac{1}{2}(w - w_0)A(w - w_0)^T.
		\end{equation}
	\end{assumption}
	Without loss of generality, we further assume that the minimum is attained at $w_0=0$, and the Hessian matrix $A$ is semi-positive definite.
	This assumption has been empirically justified in \cite{li2018visualizing}. Thus, (\ref{eq:s-fbm-sgd}) reduces to
	\begin{equation}
		\mathrm{d} w_{t}=-Aw_t \mathrm{d} t+\sigma(t, w_t) \mathrm{d}B_t^H,
		\label{eq:fou}
	\end{equation}
	the solution to which is known as a fractional Ornstein-Uhlenbeck (OU) process \cite{cheridito2003fractional}.
	For simplicity, we only consider the one-dimensional situation
	\begin{equation}
		\mathrm{d} w_{t}=-a w_t \mathrm{d} t+\sigma \mathrm{d}B_t^H,
		\label{eq:fou one dim}
	\end{equation}
	where $a$ is the corresponding diagonal entry of $A$ and $\sigma$ is a constant owing to the stationarity in the latter training stages.
	Resorting to the technique for deriving the ordinary Fokker-Planck equation for Brownian motion \cite{akhiezer2013methods, sliusarenko2011generalized}, we can obtain the state probability density of neural network weight $w$ at time $t$.
	\begin{lemma}
		\label{lemma1}
		Let $p(w, t)$ be the probability density function for  the neural network weight $w$ at time t.
		Given that $w$ follows (\ref{eq:fou one dim}) with initial condition $p(w, 0) = \delta(w-w_0)$, we explicitly have
		\begin{equation}
			p(w, t) = \frac{1}{2\sqrt{\pi\mathcal{Z}_t}}\exp\left[-\frac{\left(w - e^{-at}w_0\right)^2}{4\mathcal{Z}_t}\right],
		\end{equation}
		where
		\setlength{\arraycolsep}{0.0em}
		\begin{eqnarray}
			\mathcal{Z}_t &{}={}&\frac{a^{-2H}}{2}\sigma^2H(2H-1)\Gamma(2H-1)  \nonumber\\ 
			&&{-}\:\frac{H(2H-1)t^{2H-1}\sigma^2}{2a}E_{2H-2}(at)\nonumber\\
			&&{-}\:\frac{Ht^{2H-1}\sigma^2}{2a}e^{-2at}M(2H-1, 2H, at).\nonumber
		\end{eqnarray}
		\setlength{\arraycolsep}{5pt}
		Here, $E_p(t)$ is an integral exponential function
		\begin{equation*}
			E_p(t) = \int_{1}^{\infty} x^{-p}e^{-tx} \mathrm{~d}x,
		\end{equation*}
		and $M(a, b, t)$ is Kummer's confluent hypergeometric function
		\begin{equation*}
			\mathrm{M}(a, b, t)=\frac{\Gamma(b)}{\Gamma(a) \Gamma(b-a)} \int_{0}^{1}  \mathrm{e}^{t x} x^{a-1}(1-x)^{b-a-1} \mathrm{~d} x.
		\end{equation*}
	\end{lemma}
		To avoid clutter, we only give the proof sketch as follows and defer the complete proof to Appendix \ref{appendix: proof of lemma1}.
		\begin{IEEEproof}[Sketch of Proof]
			Assume that at time $t=0$, the neural network weight $w$ is located at $w(0)=w_{0}$. And denote by $Y_t$ the FGN corresponding to $B_t^H$. Integrating over (\ref{eq:fou one dim}), we obtain
			\begin{equation}
				w(t; w_0) = e^{-at} w_0 + \int_{0}^{t}  e^{-a(t-\tau)} Y_\tau \mathrm{~d}\tau.
				\label{eq:rawsolution}
			\end{equation}
			Moreover, the probability density function of $w$ at time $t$ is given by the Dirac delta function
			\begin{equation}
				p(w, t; w_0) = \delta\left(w - w\left(t;w_0\right)\right).
			\end{equation}
			Now that the distribution of $w_0$ is unknown, the state probability density of $w$ at any time $t$ is given by
			\begin{equation}
				p(w, t) = \int  p(w_0, 0) \left<\delta\left(w - w\left(t;w_0\right)\right) \right> \mathrm{d}w_0,
				\label{eq:intpdf}
			\end{equation}
			where $\left<\cdot\right>$ denotes an average over $Y_t$.
			Recall that
			\begin{equation}
				\delta(w) = \frac{1}{2\pi}\int  e^{iw x} \mathrm{d} x,
			\end{equation}
			and combine it with (\ref{eq:intpdf}), we obtain the analytic expression of $\hat{p}(x, t)$, which is the Fourier transform of $p(w, t)$.
			Then, according to
			\begin{equation}
				p(w, t) = \frac{1}{2\pi}\int \hat{p}(x, t)e^{iw x} \mathrm{d}x,
			\end{equation}
			we conclude the proof.
		\end{IEEEproof}
		Based on the analytical expression of $p(w, t)$, the first passage time density is obtained immediately.
		\begin{theorem}
			\label{theorem2}
			If $t>0$ is sufficiently large, the probability density function of the first passage time is approximately given by
			\begin{equation}
				p(t) \approx \mathcal{O}(\frac{a^{3H}t^{2H-2}e^{-at}}{\sigma}).
				\label{eq:escaping pdf}
			\end{equation}
		\end{theorem}
		The proof sketch is presented as follows and the complete proof is shown in Appendix \ref{appendix: proof of theorem1}.
		\begin{IEEEproof}[Sketch of Proof]
			When $t$ is sufficiently large, we will first obtain an estimation of $\mathcal{Z}(t)$ by using the limiting approximation of  $E_{2H-2}(at)$ and $M(2H-1, 2H, at)$. Following a Taylor expansion of $\exp\left[-\frac{\left(w - e^{-at}w_0\right)^2}{4\mathcal{Z}_t}\right]$, we are able to acquire an estimation of $p(w, t)$. Denote the survival probability by
			\begin{equation}
				s(t) = \int  p(w, t) \mathrm{d}w,
			\end{equation}
			then, it is direct to obtain the first passage density
			\begin{equation}
				p(t)=-\frac{\mathrm{d} s(t)}{\mathrm{d}t} \approx \mathcal{O}(\frac{a^{3H}t^{2H-2}e^{-at}}{\sigma}).
				\label{eq:firstpassage time dnesity}
			\end{equation}
		\end{IEEEproof}
	\begin{remark}
		It should be noted that in previous works \cite{sliusarenko2010kramers, sanders2012first} motion correlations at different moments were neglected, and a pseudo-Markovian scheme was utilized to approximate the state probability density and obtain a similar result $p(t) \approx \mathcal{O}(t^{2H-2})$.
		However, for the case $H>1/2$, the integral for computing the mean passage time diverges. In numerical simulations, this can be circumvented by using Feynman's path integral, as suggested in \cite{jumarie1993stochastic}.
		By contrast, the proposed approximation provides a remedy and includes the dependence on local curvature $a$ and noise intensity $\sigma$.
	\end{remark}
	\begin{remark}
		Since the constant $a$ indicates the curvature at the local minimum; thus, by integrating \eqref{eq:escaping pdf} over time $t$, it is easy to verify that:
		(1) it requires more time to exit for large $a$ and $H$; (2) it requires less time to exit for large $\sigma$.
		This observation is empirically confirmed as well, see Fig. \ref{fig:escaping_time}.
		This implies that at early training stages, on one hand, larger noise helps SGD effectively escape from sharp minima, while on the other hand, at final stages, larger $H$ enforces SGD to stay longer in flat, or so-called wide, minima. 
		This serves as an evident explanation that SGD prefers wide minima that generalize well \cite{simsekli2019tail, nguyen2019first}.
	\end{remark}
	\begin{figure*}[t]
		\centering
		\includegraphics[width=1.0\linewidth]{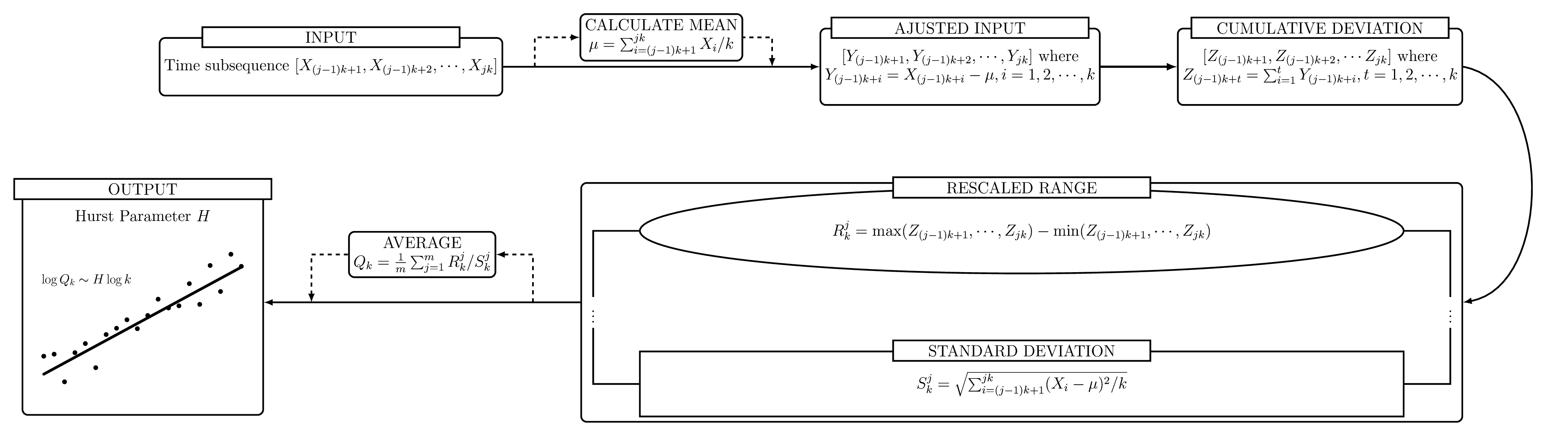}
		\caption{Diagram for $R/S$ analysis. Given a time series of length $N$, we first split it into $m$ subsequences of length $k$. Then, for each subsequence, we calculate its $R/S$ statistic and return the average $Q_k$. Using a simple linear fitting on the log-log plot, we obtain an estimate of the Hurst parameter $H$, which is the slope.}
		\label{fig:hurst flowchart}
	\end{figure*}
	\section{Experiments}
	In this section, we conduct extensive experiments to support our conjecture that SGN follows FBM instead of Brownian motion or L\'evy random process:
	\begin{enumerate}
		\item We first argue that SGN is neither a Gaussian nor a L\'evy stable distribution using existing statistical testing methods;
		\item Subsequently, we simulate a simple high-dimensional SDE  to verify that the convergence diversity is caused by different fractional noises and that the Euclidean distance from the initialization of network weights $w$ correlates well with the Hurst parameter;
		\item In the third part, we study the mean escaping time of the neural parameter $w$ trapped in a harmonic potential to demonstrate that a larger Hurst parameter yields a lower escaping rate. Then, we approximately compute the largest eigenvalue of the Hessian matrix for shallow neural networks to verify that a large Hurst parameter corresponds to flat minima;
		\item Finally, we investigate the short-range memory effects on a broad range of model architectures,  datasets, and training strategies.
	\end{enumerate}
	\par
	\emph{\textbf{Datasets}}. The datasets we use in the experiments are MNIST, CIFAR10, and CIFAR100. We randomly split the MNIST dataset into training and testing subsets of sizes 60K and 10K, and the CIFAR10 and CIFAR100 datasets into training and testing subsets of sizes 50K and 10K, respectively.
	Several types of neural networks are utilized, including a fully connected network (FCN) and several convolutional neural networks (CNNs). 
	The total model parameters scale from thousands to millions, as is the case in most DNNs.
	\par
	\emph{\textbf{Implementation Details}}. We run every single model three times with different random seeds, and the mean is plotted throughout the experiments. To train the model parameters, we select the vanilla SGD without additional techniques such as momentum or weight decaying. Without any specification, we use a batch size of 64 and a learning rate of 0.01. Similar to large-scale studies \cite{jiang2020fantastic, dziugaite2020search}, we terminate the training process once the cross-entropy loss decreases to a threshold of 0.01. And we set the same batch size as in the training process to estimate the SGN vectors. To reproduce the results, the code is available at \url{https://github.com/cltan023/shortRangeSGD}.
		\par
\begin{figure}[t]
			\centering
			\begin{subfigure}[b]{0.24\textwidth}
				\centering
				\includegraphics[width=\textwidth, clip, trim= 0 0 0 0]{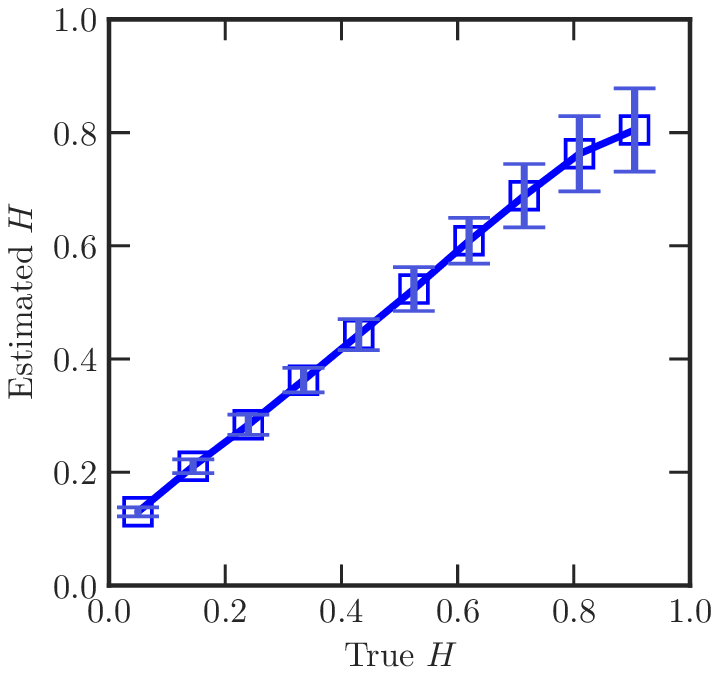}
				\caption{}
			\end{subfigure}
			\begin{subfigure}[b]{0.24\textwidth}
				\centering
				\includegraphics[width=\textwidth, clip, trim= 0 0 0 0]{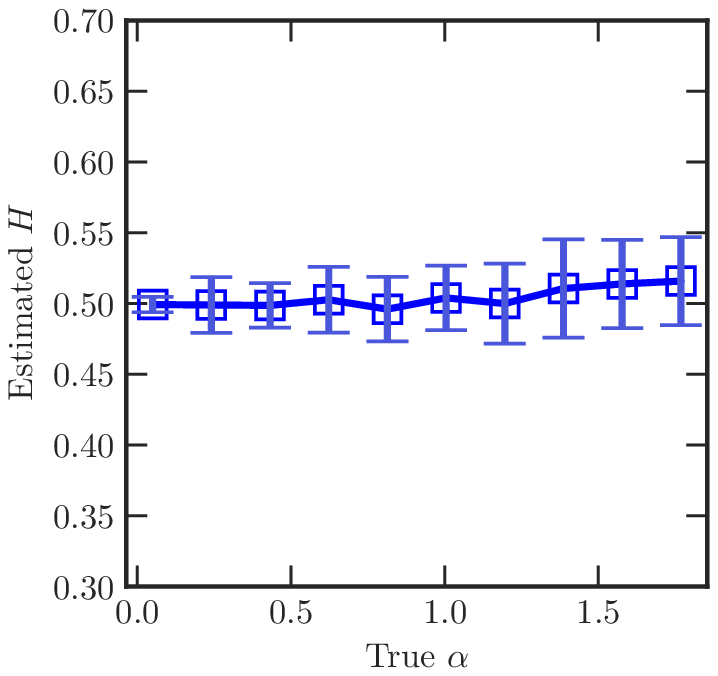}
				\caption{}
			\end{subfigure}
			\caption{Hurst parameter estimation for (a) FBM and (b) L\'evy $\alpha$-stable distribution.}
			\label{fig:hurst examine}
		\end{figure}

		\emph{\textbf{Measurement Metric}}. We now introduce the estimation method for the Hurst parameter. There are several such techniques. However, as real-time series are finite in practice, and the local dynamics corresponding to a particular temporal window can be over-estimated, no estimator can be universally effective. Established approaches to estimating the Hurst parameter can be classified into three categories. They are applied either in the time domain, such as detrended fluctuation and rescaled range analysis or in the frequency domain, such as the wavelet and the periodogram method. Alternatively, a radically different method called visibility graph was proposed in \cite{lacasa2009visibility}, whereby the time series is mapped onto a network using a visibility algorithm. An overview of related work can be found in \cite{embrechts2009selfsimilar}. In this study, we use the classical rescaled range ($R/S$) method \cite{qian2004hurst} for simplicity and efficiency.
		\begin{definition}
				Let $\{X_i\}_{i=1}^k$ be a sequence of random variables and the mean and standard deviation are given by
				\begin{equation}
					\mu = \frac{1}{k}\sum_{i=1}^{k}X_i, \quad S_k = \sqrt{\frac{1}{k}\sum_{i=1}^{k}(X_i - \mu)^2}.
				\end{equation}
				We define another two sequences $\{Y_i\}_{i=1}^k$ and $\{Z_t\}_{t=1}^k$ respectively by,  
				\begin{equation}
					Y_i = X_i -\mu, i = 1, 2, \cdots, k
				\end{equation}
				and 
				\begin{equation}
					Z_t = \sum_{i=1}^{t} Y_i, t = 1, 2, \cdots, k.
				\end{equation}
				Let
				\begin{equation}
					R_k = \max(Z_1, Z_2, \cdots, Z_k) - \min(Z_1, Z_2, \cdots, Z_k),
				\end{equation}
				the ratio $Q_k=R_k/S_k$
				is then called the rescaled adjusted range or $R/S$ statistic.
		\end{definition}
		To use the $R/S$ statistic for the estimation of the Hurst parameter, we first recall a theorem from \cite{embrechts2009selfsimilar}.
		\begin{theorem}
			Let $\{X_t\}$ be a strictly stationary stochastic process such that $\{X^2_t\}$ is ergodic, and $k^{-H} \sum_{j=1}^{[k t]} X_j$ converges weakly to FBM as $k$ tends to infinity. Then, 
			\begin{equation}
				k^{-H} Q_k \stackrel{\text { dist. }}{\longrightarrow} \xi,~\text{as}~k\to\infty, 
			\end{equation}
			where $\xi$ is a non-degenerate random variable and $\stackrel{\text { dist. }}{\longrightarrow}$ means convergence in distribution.
		\end{theorem}
			From the perspective of statistical analysis, this theorem implies that for a  sufficiently large $k$, the points in the plot of $\log Q_k$ against $\log k$ will be randomly scattered around a straight line with slope $H$.
			Practically, to estimate the Hurst parameter of a time series $\{X_i\}$ of length $N$, the procedure generally works as follows:
			\begin{enumerate}
				\item choose a set of $k$ such that $N$ is divisible by $k$;
				\item for each $k$, split the time series $\{X_i\}_{i=1}^N$ into $m$ subsequences $[X_1, X_2, \ldots, X_k]$, $[X_{k+1}, X_{k+2}, \ldots, X_{2k}]$, $\cdots$, $[X_{N-k+1}, X_{N-k+2}, X_N]$, where $m = \lfloor N/k \rfloor$. For each subsequence, calculate the statistics $R_k^j$ and $S_k^j$ according to Definition 4 and return the average $Q_k=\frac{1}{m}\sum_{j=1}^{m}R_k^j/S_k^j$;
				\item fit the pairs $\{\log k, \log Q_k\}$ with a linear function and the resultant slope is the so-called Hurst parameter $H$.
			\end{enumerate}
		The detailed steps for estimating the Hurst parameter for a time series are summarized in Fig. \ref{fig:hurst flowchart}.
		To determine the efficiency of the $R/S$ method, we first uniformly sample 100 different values of $H$ from the interval $(0, 1)$. Then, for each $H$, we generate 1000 trajectories of the FBM series consisting of 10000 points. The averages and standard deviations are shown in Fig. \ref{fig:hurst examine}(a).
		Although $H$ is underestimated when the true Hurst parameter is smaller than 0.2 or larger than 0.8, the estimation is accurate in most cases.
		\subsection{SGN is neither Gaussian nor L\'evy Stable}
\begin{figure}[th!]
			\centering
			\begin{subfigure}[b]{0.24\textwidth}
				\centering
				\includegraphics[width=\textwidth, clip, trim= 0 0 0 0]{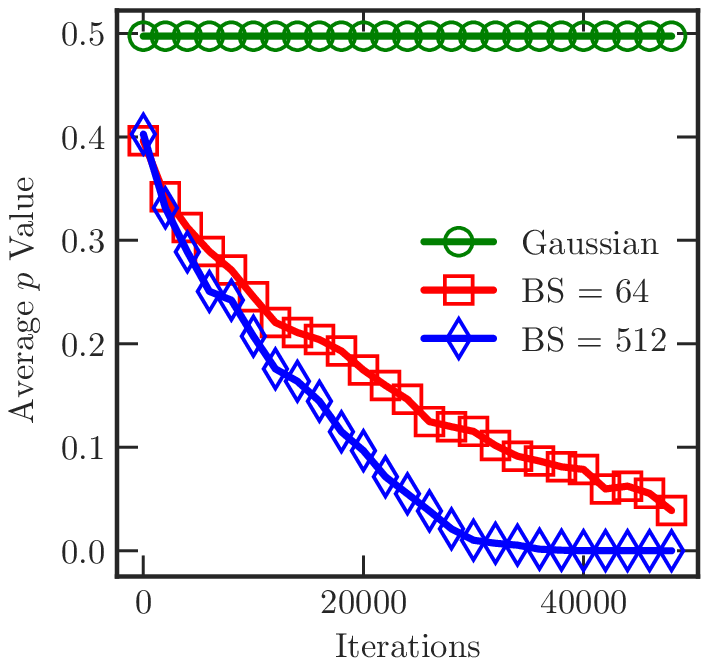}
				\caption{}
			\end{subfigure}
			\begin{subfigure}[b]{0.24\textwidth}
				\centering
				\includegraphics[width=\textwidth, clip, trim= 0 0 0 0]{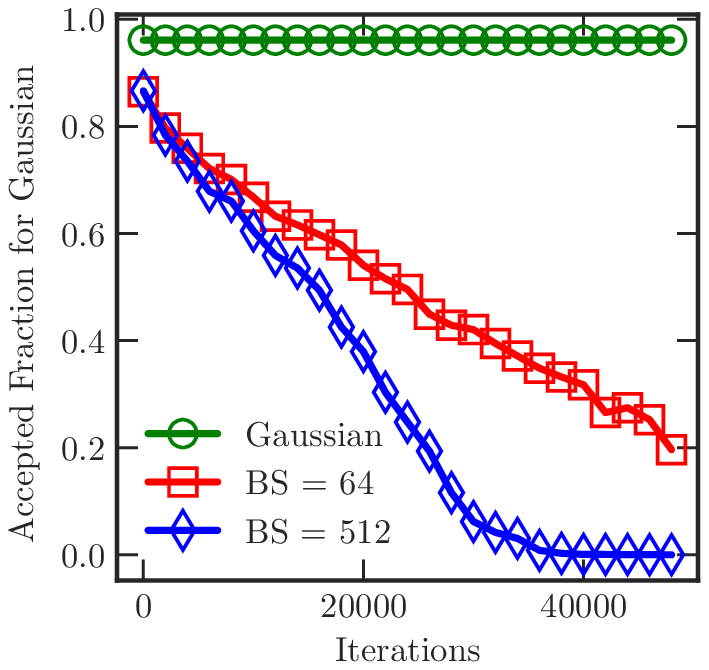}
				\caption{}
			\end{subfigure}
			\caption{(a) Shapiro-Wilk and (b) Anderson-Darling Gaussianity test result for ResNet18 on CIFAR10. A smaller value implies that the SGN is less likely to be Gaussian.}
			\label{fig:gaussian test}
		\end{figure}

		We first use the classical Shapiro-Wilk and Anderson-Darling test methods suggested by \cite{panigrahi2019non} to determine the Gaussianity of SGN.
		We train a ResNet18 network on CIFAR10 for batch sizes of 64 and 512. Then, we log the SGN vectors at every 100th iteration and conduct the Shapiro-Wilk and Anderson-Darling tests. As shown in Fig. \ref{fig:gaussian test}, the SGN becomes highly non-Gaussian as the training continues. More experimental results can be found in \cite{panigrahi2019non}.
		\par
		In fact, as SGN has been shown to be non-Gaussian in \cite{panigrahi2019non}, we are more interested in determining whether or not SGN follows a L\'evy stable distribution. To this end, we utilize another result from \cite{embrechts2009selfsimilar}.
		\begin{theorem}
			Let $\{X_t\}$ be i.i.d. random variables with $\mathbb{E}[X_t^2]=\infty$ in the domain of L\'evy stable distribution. Then,
			\begin{equation}
				k^{-1/2} Q_k \stackrel{\text { dist. }}{\longrightarrow} \xi, ~\text{as}~k\to\infty
			\end{equation}
			where $\xi$ is a non-degenerate random variable and $\stackrel{\text { dist. }}{\longrightarrow}$ means convergence in distribution.
		\end{theorem}	
			This theorem implies that if the SGN follows a L\'evy stable distribution, then estimations of the Hurst parameter in the log-log plot are scattered around a straight line with a slope of $0.5$.
			Therefore, the reasoning line that SGN does not follow a L\'evy stable distribution becomes clear:
			\begin{enumerate}
				\item verify that if samples are generated from L\'evy stable distributions, then the estimated Hurst parameter is 0.5;
				\item estimate the Hurst parameter of SGN, and if the estimated Hurst parameter is far different from 0.5, then SGN does not follow the L\'evy stable distribution.
			\end{enumerate}
			To validate 1), we first uniformly sample 100 different values of $\alpha$ from the interval $(0, 2)$. 
			Then, for each $\alpha$, we generate 1000 sequences of L\'evy stable series with 10000 points.
			Fig. \ref{fig:hurst examine}(b) shows the averages and standard deviations.
			It can be seen that the estimated Hurst parameter is not sensitive to $\alpha$ and remains unchanged around the horizontal line of $H=0.5$. This implies that noise generated from a L\'evy stable distribution exhibits no sign of short-range memory effects.
		By contrast, the Hurst estimation of the SGN from, for instance, Figs. \ref{fig:fcn_mnist} and  \ref{fig:alexnet_cifar10_100}, is far away from 0.5 and varies according to different hyperparameters. 
		Therefore, it is reasonable to claim that SGN does not follow a L\'evy stable distribution.
		\subsection{Larger Hurst Parameter Indicates Faster Convergence}
		The convergence rate is of great interest in the study of SDEs. Theorem \ref{theorem1} implies that the convergence rate of FBM-driven SDEs is positively correlated with the Hurst parameter. Herein, we experimentally confirm this theoretical observation and provide further evidence that the SGN has short-range memory.
		\par
\begin{figure}[t]
			\centering
			\begin{subfigure}[b]{0.24\textwidth}
				\centering
				\includegraphics[width=\textwidth, clip, trim= 0 0 0 0]{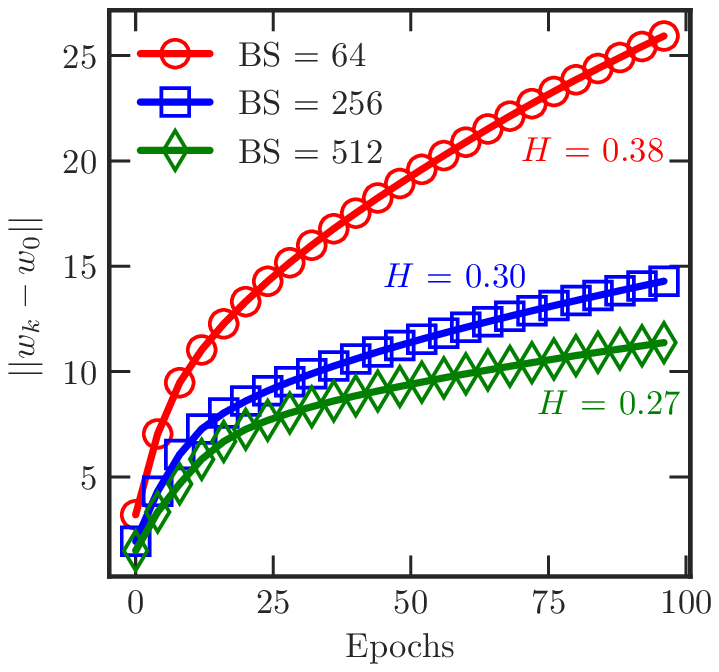}
				\caption{}
			\end{subfigure}
			\begin{subfigure}[b]{0.24\textwidth}
				\centering
				\includegraphics[width=\textwidth, clip, trim= 0 0 0 0]{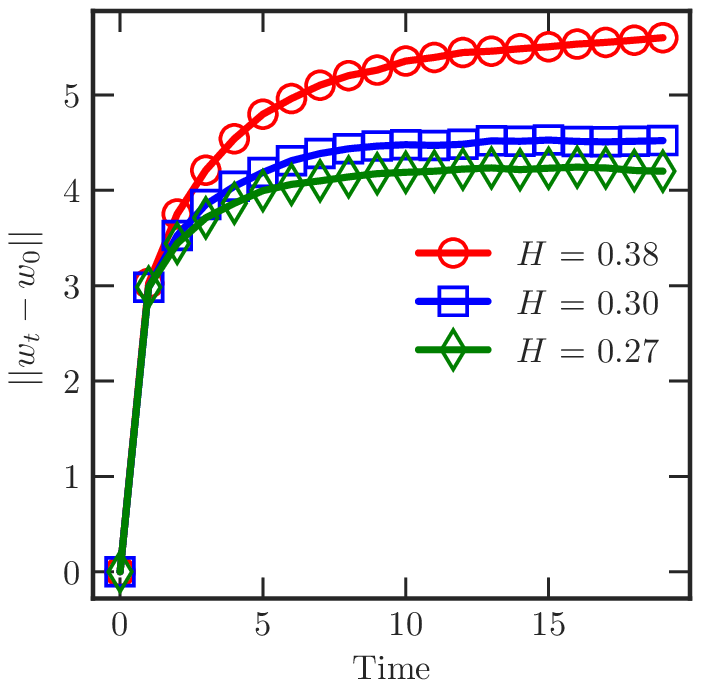}
				\caption{}
			\end{subfigure}
			\caption{(a) Euclidean distance of neural network weights $w$ from initialization for different batch sizes, and the corresponding estimations of Hurst parameter. (b) Euclidean distance of the simulation of SDE from the origin driven by the same Hurst parameters.}
			\label{fig:resnet18_cifar10_moving_distance}
		\end{figure}

			It is well-recognized that large noise intensity in small batch size encourages the neural weights $w$ to go further from the initialization \cite{keskar2017large, hoffer2017train}. 
			In this section, we provide an alternative explanation and argue that it may result from different Hurst parameters.
			We first train several ResNet18 models on CIFAR10 with different batch sizes and estimate the corresponding Hurst parameters. 
			It should be highlighted that at each step of training, the noise intensity of all models is tuned to the same magnitude, to cancel the effect of noise intensity.
			As shown in Fig. \ref{fig:resnet18_cifar10_moving_distance} (a), the Euclidean distance of neural network weight $w$ from the initialization still decreases with the batch size. 
			\par
			Next, we apply the estimated Hurst parameters to the following SDE
			\begin{equation}
				\mathrm{d}{w_t} = - 2w_t\mathrm{d}t + \sigma \mathrm{d}B_t^H,
			\end{equation}
			where $w_t\in \mathbf{R}^{d}$, and $\sigma$ is a constant. 
			Without loss of generality, we assume that $d=100000$ and $\sigma=0.01$.
			With the initial condition that $w=\mathbf{0}$, we are interested in the Euclidean distance from the origin for different Hurst parameters as time evolves.
			As shown in Fig. \ref{fig:resnet18_cifar10_moving_distance} (b), when the Hurst parameter is larger, the solution converges at a faster rate and drifts further away from the initialization.
			This is in line with Theorem \ref{theorem1}; therefore,  except for the noise intensity, the Hurst parameter is another major source of convergence diversity for different batch sizes.
		\begin{figure}[t]
			\centering
			\includegraphics[width=0.48\textwidth, clip, trim= 0 0 0 0]{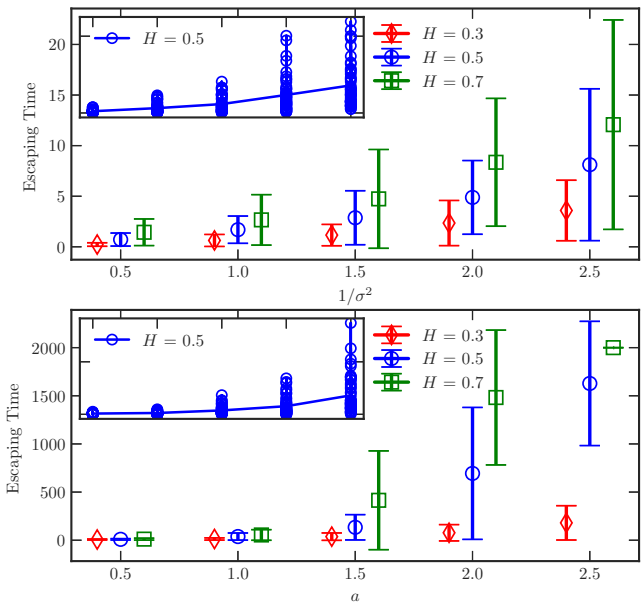}
			\caption{
				Influence of $\sigma$ (upper panel) and $a$ (lower panel) on the mean escaping time of FBM-driven SDE at $H=0.3$, $H=0.5$ and $H=0.7$. 
				For $a=2.5$ and $H=0.7$, it is noteworthy that $w$ fails to reach the absorbing boundary when the pre-specified time expires. Thus, all points are truncated to concentrate on a single value.The inset shows all the escaping times at $H=0.5$ in detail, and the line connects the means.}
			\label{fig:escaping_time}
		\end{figure}
		\subsection{Larger Hurst Parameter Indicates Wider Minima}
		In this section, we study the escaping behavior from local minima.
		For simplicity, we only consider one-dimensional situation in which the neural network weight $w$ is trapped in a local minimum at $w=0$.
		We assume that the loss function around this point is convex and can be written as 
		\begin{equation}
			f(w) = \frac{a}{2}w^2.
		\end{equation}
		Thus, the dynamics in the vicinity of the local minimum can be described as
		\begin{equation}
			\mathrm{d}w_t = -aw_t \mathrm{d}t + \sigma \mathrm{d}B_t^H.
		\end{equation}
		Without loss of generality, given an absorbing boundary $w=\pm1$, we are concerned with the mean time so that $w$ can reach this boundary. For this purpose, we select a time step $\eta=0.01$, and simulate 50 times for different Hurst parameters $H=0.3$, $H=0.5$, and $H=0.7$.
		To study the influence of local curvature and noise intensity, we select $a$ and the reciprocal of $\sigma^2$ both from the range $\{0.5, 1.0, 1.5, 2.0, 2.5\}$. 
		For each run, we first generate a sequence of $10^5$ FGN series as the noise input.
		As can be seen from Fig. \ref{fig:escaping_time}, the time for $w$ to escape from the harmonic potential grows with the Hurst parameter. Furthermore, it is heavily affected by the local curvature and the noise intensity. It shows that it is difficult to escape from a sharp bowl with a low noise intensity.
		\par
\begin{figure}[t]
			\centering
			\begin{subfigure}[b]{0.24\textwidth}
				\centering
				\includegraphics[width=\textwidth, clip, trim= 0 0 0 0]{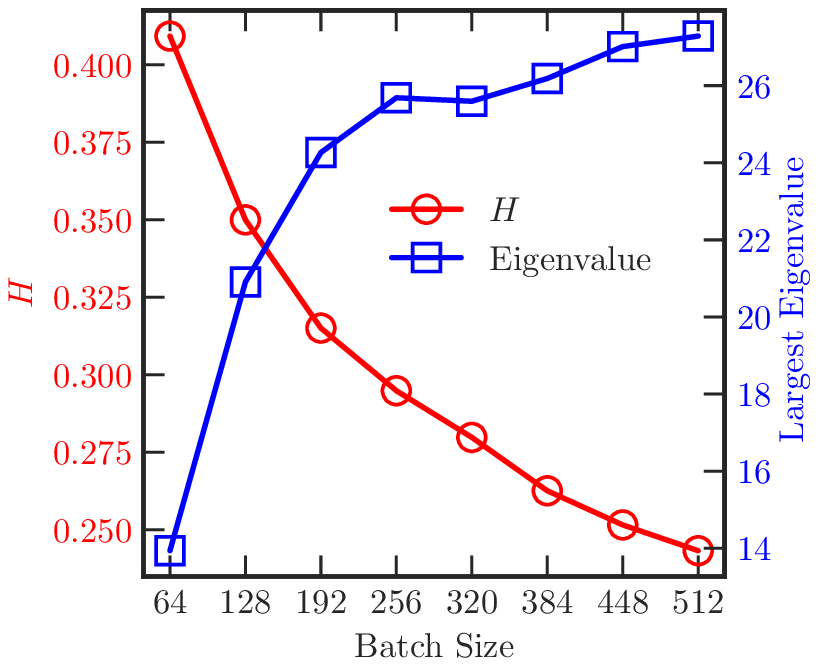}
				\caption{}
			\end{subfigure}
			\begin{subfigure}[b]{0.24\textwidth}
				\centering
				\includegraphics[width=\textwidth, clip, trim= 0 0 0 0]{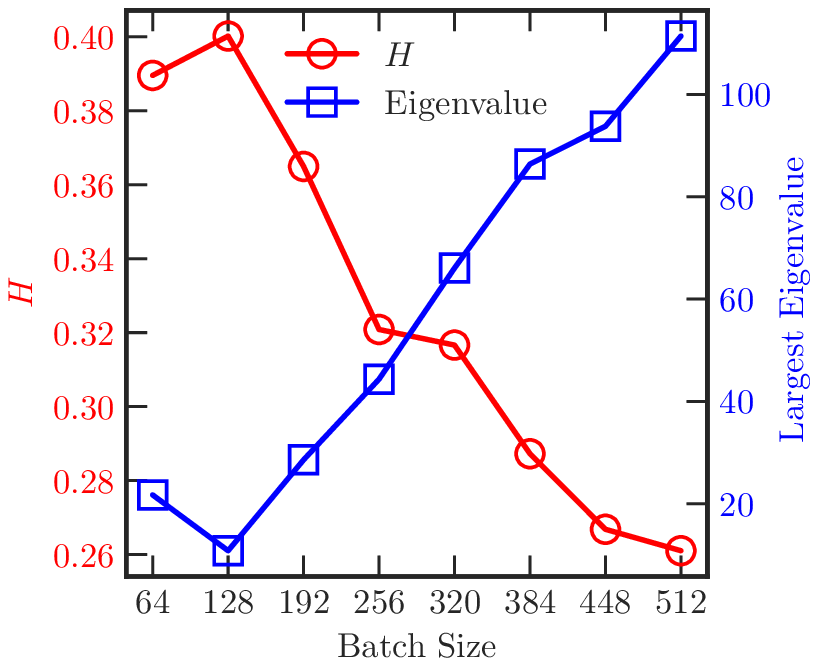}
				\caption{}
			\end{subfigure}
			\caption{Estimation of the Hurst parameter and largest eigenvalue for different batch sizes: (a) FCN on MNIST; (b) AlexNet on CIFAR10.}
			\label{fig:fc_alexnet_eigen}
		\end{figure}

		We now empirically confirm the assertion that larger Hurst parameter leads to wider minima. To this end, we select a single FCN and AlexNet as the base network. We train the FCN network on MNIST, and AlexNet on CIFAR10.
		The batch size is varied in $\{64, 128, \cdots, 512\}$. Similar to \cite{chaudhari2019entropy}, we compute the largest eigenvalue of the Hessian matrix at the last iteration as a measure of flatness.
		Large eigenvalues generally indicate sharp minima.
		Fig. \ref{fig:fc_alexnet_eigen} shows that the largest eigenvalue increase with the batch size, whereas the Hurst parameter changes in the opposite direction.
		This implies that a small Hurst parameter results in large eigenvalues and tends to end in sharp minima. 
		\subsection{Impacts of Hyperparameters}
\begin{figure}[t]
			\centering
			\begin{subfigure}[b]{0.24\textwidth}
				\centering
				\includegraphics[width=\textwidth, clip, trim= 0 0 0 0]{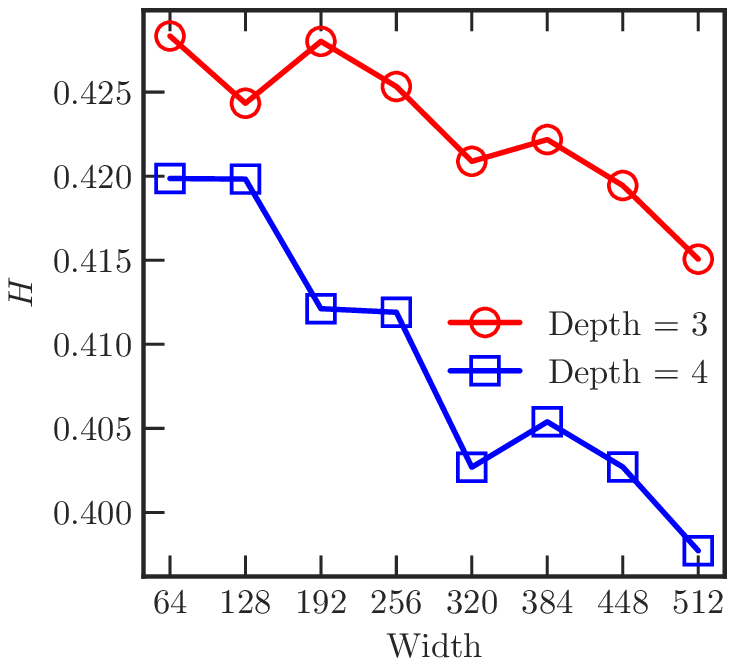}
				\caption{}
			\end{subfigure}
			\begin{subfigure}[b]{0.24\textwidth}
				\centering
				\includegraphics[width=\textwidth, clip, trim= 0 0 0 0]{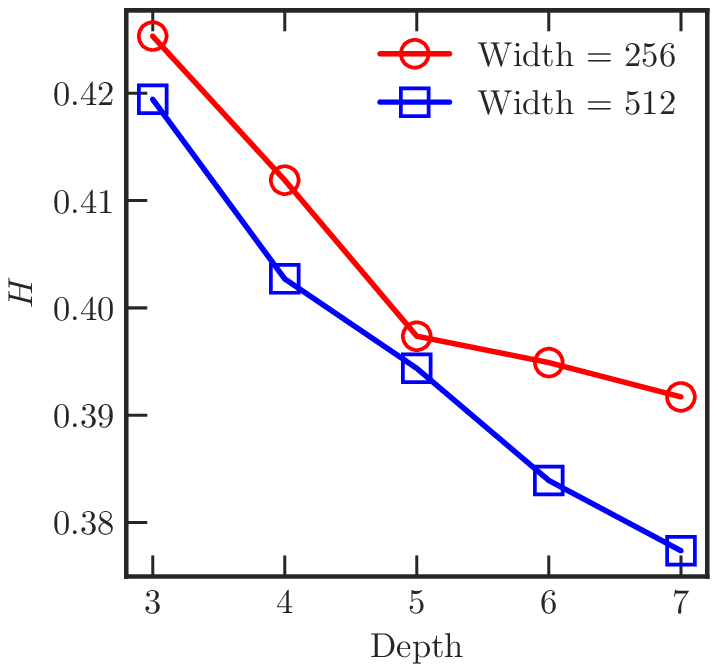}
				\caption{}
			\end{subfigure}
			\caption{Estimation of the Hurst parameter for different widths and depths for FCN networks on MNIST.}
			\label{fig:fcn_mnist}
		\end{figure}

\begin{figure}[t]
			\centering
			\begin{subfigure}[b]{0.24\textwidth}
				\centering
				\includegraphics[width=\textwidth, clip, trim= 0 0 0 0]{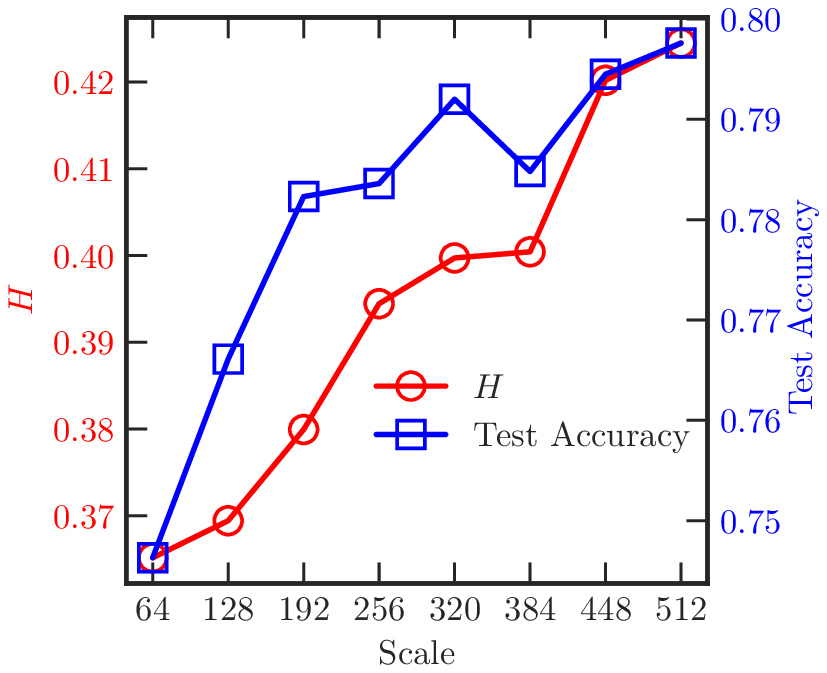}
				\caption{}
			\end{subfigure}
			\begin{subfigure}[b]{0.24\textwidth}
				\centering
				\includegraphics[width=\textwidth, clip, trim= 0 0 0 0]{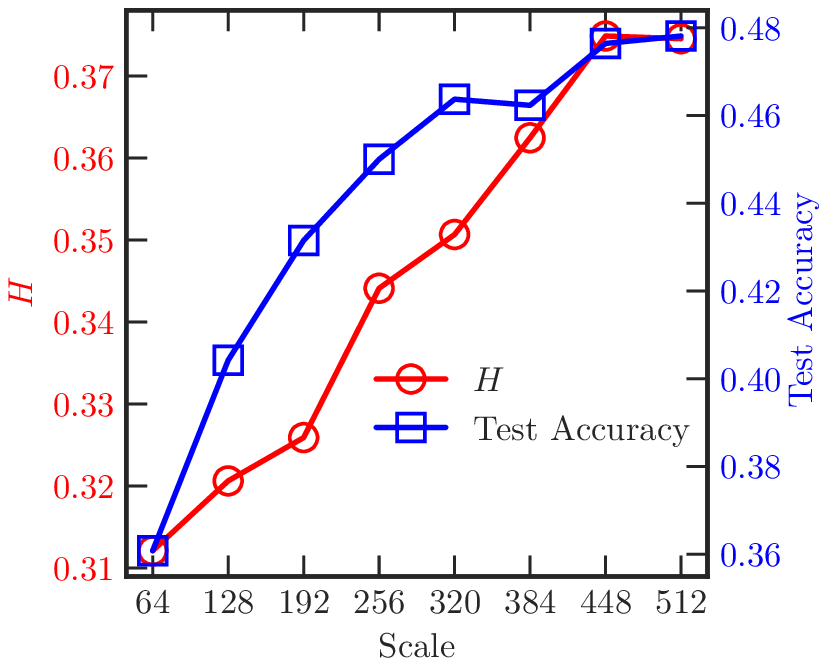}
				\caption{}
			\end{subfigure}
			\caption{Estimation of the Hurst parameter and testing accuracy across varying scales for AlexNet on (a) CIFAR10 and (b) CIFAR100.}
			\label{fig:alexnet_cifar10_100}
		\end{figure}

		\emph{\textbf{Impacts of model capacity.}} We first investigate the influence of model capacity on the Hurst parameter.
		For simple FCN networks, we vary the width (number of neurons in each layer) and depth (number of hidden layers).
		By contrast, for complex CNN networks, we alter the scale (number of feature maps) in each convolution layer.
		For FCN networks, we select the width from $\{64, 128, \cdots, 512\}$ and the depth from $\{3, 4, \cdots, 7\}$, and train $40$ models in total on the MNIST dataset in each run. As can be seen from Fig. \ref{fig:fcn_mnist}, the Hurst parameter is sensitive to the model architecture and decreases with the width and depth of the FCN. As to CNN networks, we select AlexNet as the base model and range the filters in each layer from $\{64, 128, \cdots, 512\}$.
		Fig. \ref{fig:alexnet_cifar10_100} shows that the Hurst parameter is highly sensitive to network scale, and monotonically increases with model capacity on  CIFAR10 and CIFAR100.
		
		\begin{figure}[t]
			\centering
			\begin{subfigure}[b]{0.24\textwidth}
				\centering
				\includegraphics[width=\textwidth, clip, trim= 0 0 0 0]{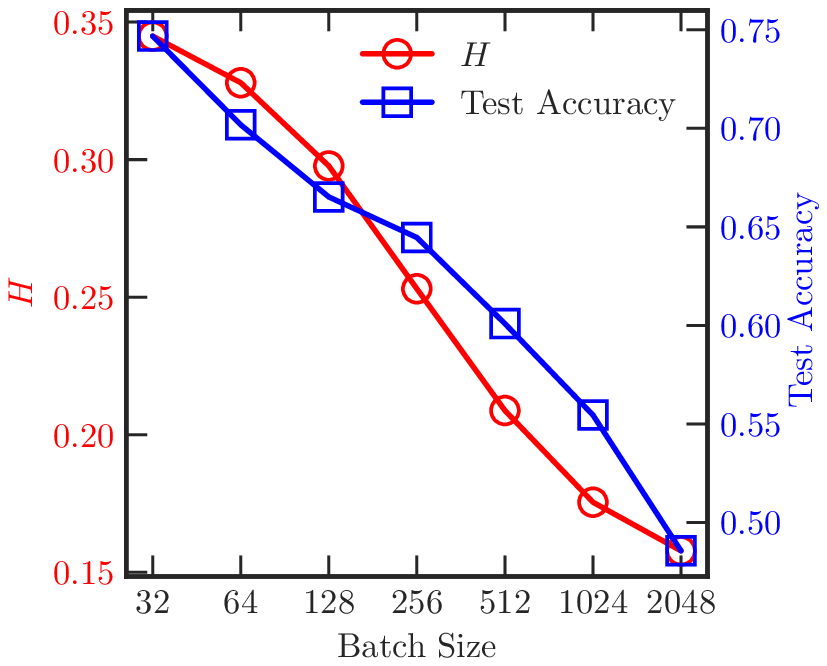}
				\caption{}
			\end{subfigure}
			\begin{subfigure}[b]{0.24\textwidth}
				\centering
				\includegraphics[width=\textwidth, clip, trim= 0 0 0 0]{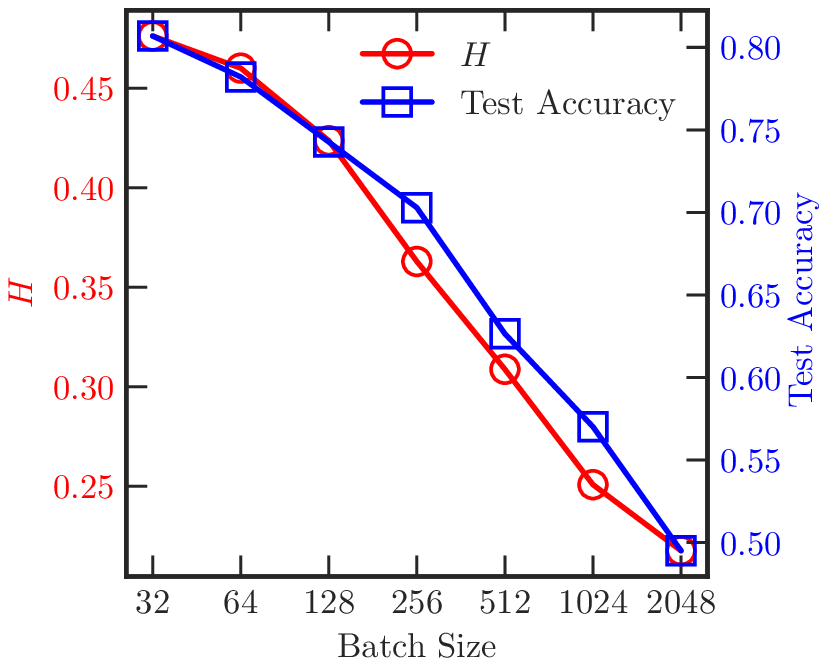}
				\caption{}
			\end{subfigure}
			\caption{Estimation of the Hurst parameter and test accuracy across varying batch sizes for (a) ResNet18 and (b)VGG16\_BN on CIFAR10.}
			\label{fig:resnet_cifar10}
		\end{figure}
		
		\begin{figure}[h!]
			\centering
			\begin{subfigure}[b]{0.24\textwidth}
				\centering
				\includegraphics[width=\textwidth, clip, trim= 0 0 0 0]{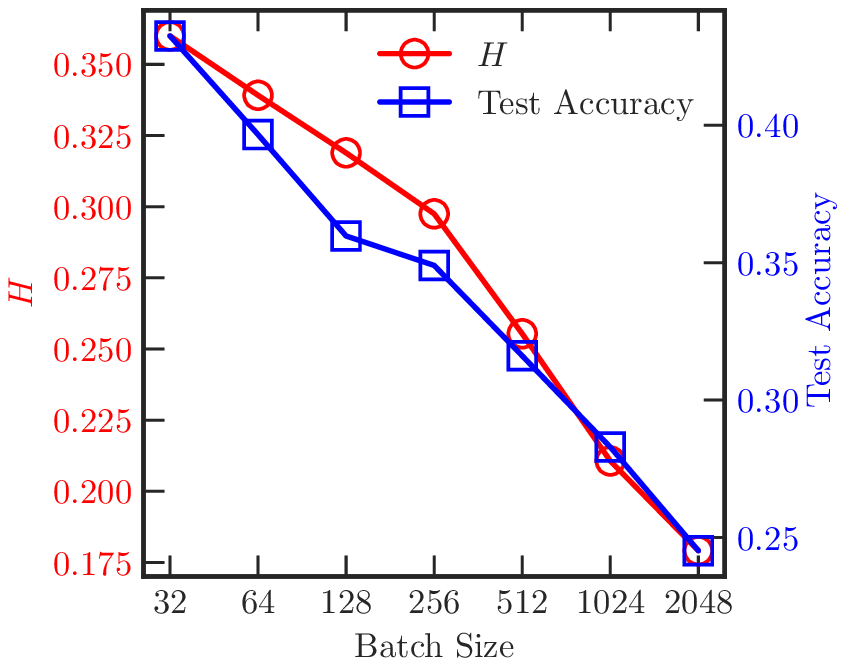}
				\caption{}
			\end{subfigure}
			\begin{subfigure}[b]{0.24\textwidth}
				\centering
				\includegraphics[width=\textwidth, clip, trim= 0 0 0 0]{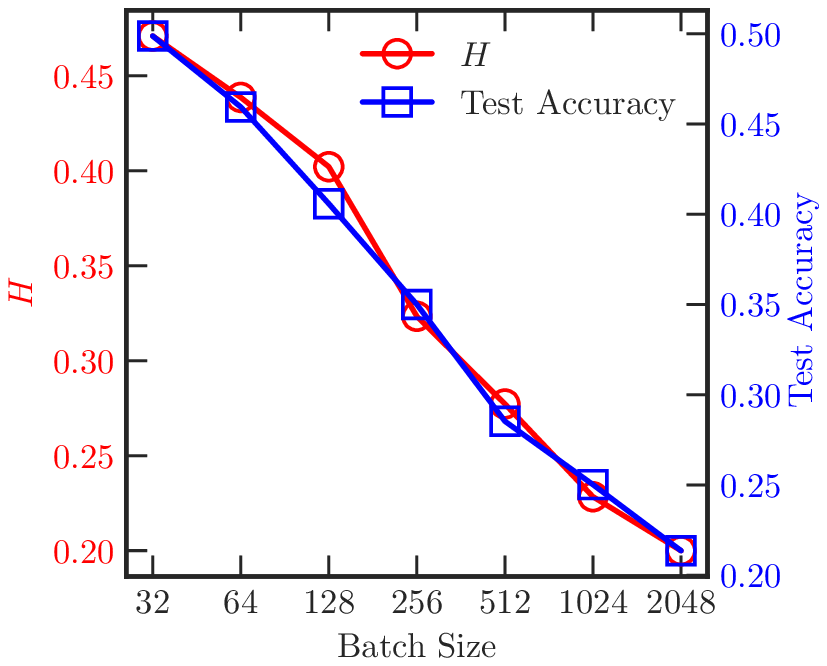}
				\caption{}
			\end{subfigure}
			\caption{Estimation of the Hurst parameter and test accuracy across varying batch sizes for (a) ResNet18 and (b)VGG16\_BN on CIFAR100.}
			\label{fig:resnet_cifar100}
		\end{figure}
		\par
		\emph{\textbf{Impacts of batch size and learning rate}.} It is well-known that a large batch size generally leads to sharp minima and thus poor generalization. Herein, we examine the effects of batch size on the Hurst parameter.
		According to the discussion in the previous section, it is reasonable to conclude that the Hurst parameter decreases with batch size.
		To further verify this observation, we monitor the behavior of the Hurst parameters for ResNet18 and VGG16\_BN on CIFAR10 and CIFAR100.
		The law that the Hurst parameter decreases with the batch size is preserved, as shown in Figs. \ref{fig:resnet_cifar10} and  \ref{fig:resnet_cifar100}.
		\par
\begin{figure}[t]
			\centering
			\begin{subfigure}[b]{0.24\textwidth}
				\centering
				\includegraphics[width=\textwidth, clip, trim= 0 0 0 0]{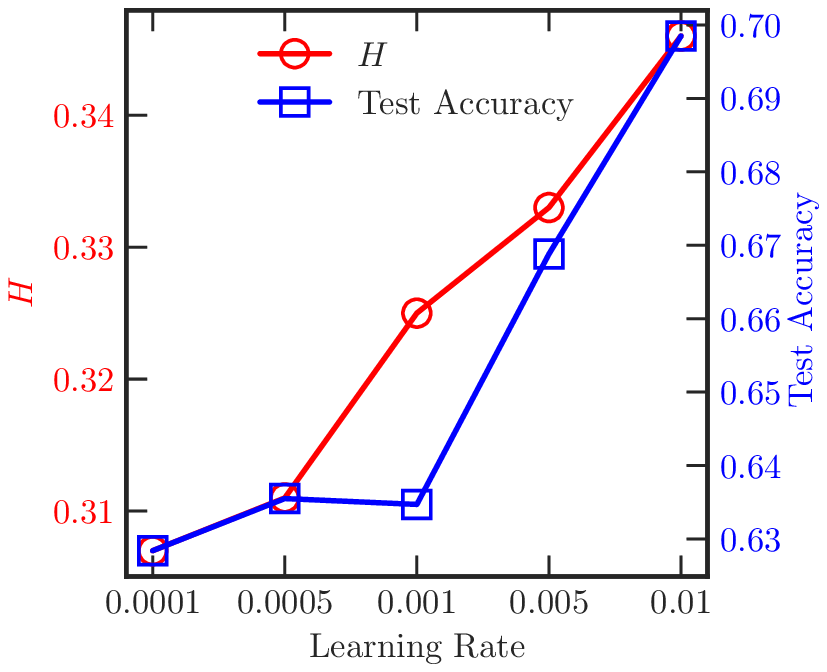}
				\caption{}
			\end{subfigure}
			\begin{subfigure}[b]{0.24\textwidth}
				\centering
				\includegraphics[width=\textwidth, clip, trim= 0 0 0 0]{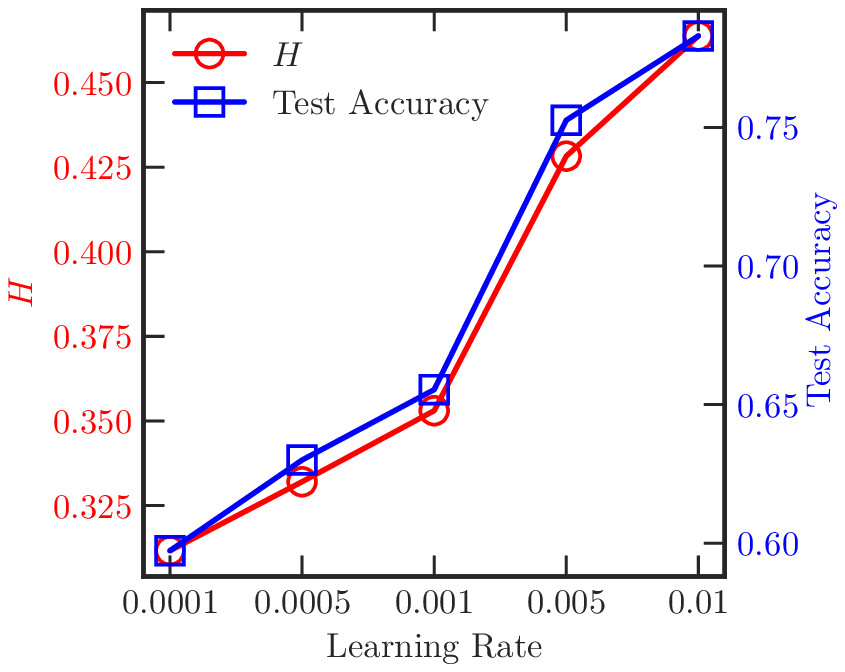}
				\caption{}
			\end{subfigure}
			\caption{Estimation of the Hurst parameter and test accuracy across varying learning rates for (a) ResNet18 and (b)VGG16\_BN on CIFAR10.}
			\label{fig:influence_of_learning_rate_cifar10}
		\end{figure}

		We also investigate the influence of the learning rate on the Hurst parameter for ResNet18 and VGG16\_BN. 
		The learning rate is selected from $\{0.0001, 0.0005, 0.001, 0.005, 0.01\}$.
		As can be seen from Fig. \ref{fig:influence_of_learning_rate_cifar10}, the Hurst parameter increases with the learning rate; this is aligned with the empirical observation that a large learning rate leads to better generalization \cite{he2019control}.
		
		\begin{figure}[t]
			\centering
			\begin{subfigure}[b]{0.24\textwidth}
				\centering
				\includegraphics[width=\textwidth, clip, trim= 0 0 0 0]{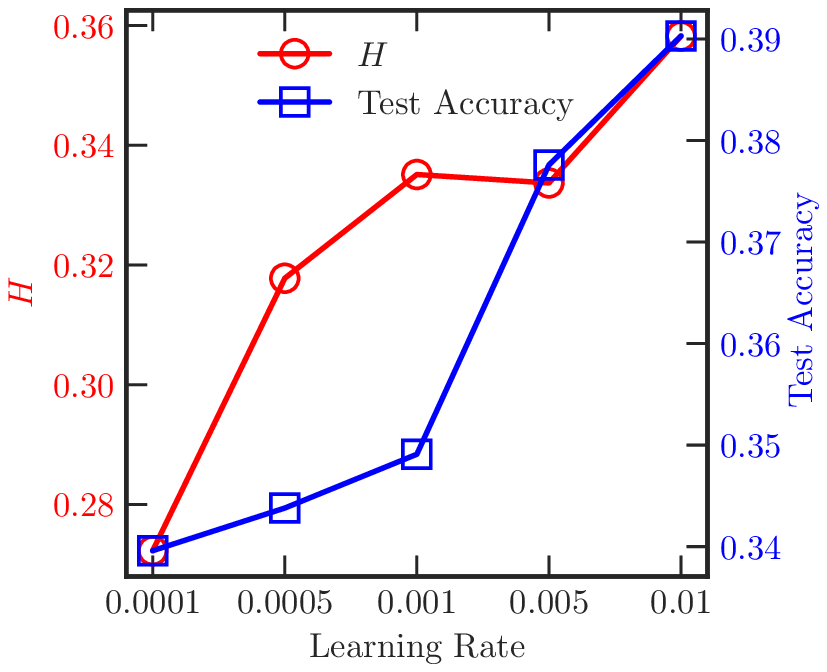}
				\caption{}
			\end{subfigure}
			\begin{subfigure}[b]{0.24\textwidth}
				\centering
				\includegraphics[width=\textwidth, clip, trim= 0 0 0 0]{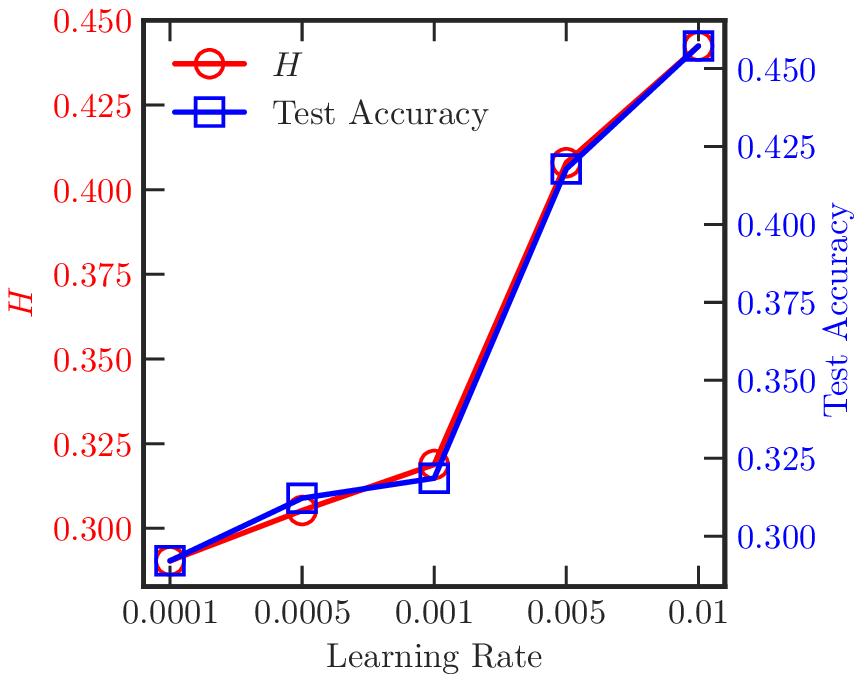}
				\caption{}
			\end{subfigure}
			\caption{Estimation of the Hurst parameter and test accuracy across varying learning rates for (a) ResNet18 and (b)VGG16\_BN on CIFAR100.}
			\label{fig:influence_of_learning_rate_cifar100}
		\end{figure}
		\begin{figure}[h!]
			\centering
			\begin{subfigure}[b]{0.24\textwidth}
				\centering
				\includegraphics[width=\textwidth, clip, trim= 0 0 0 0]{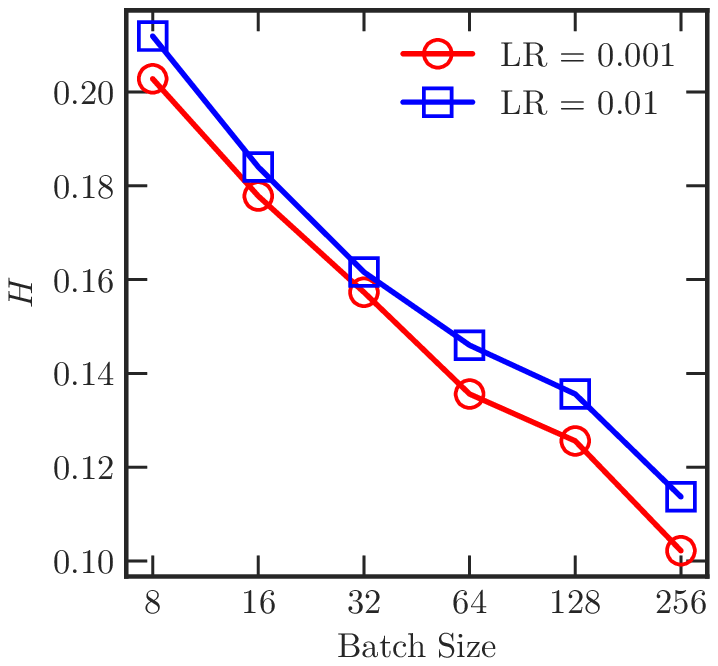}
				\caption{}
			\end{subfigure}
			\begin{subfigure}[b]{0.24\textwidth}
				\centering
				\includegraphics[width=\textwidth, clip, trim= 0 0 0 0]{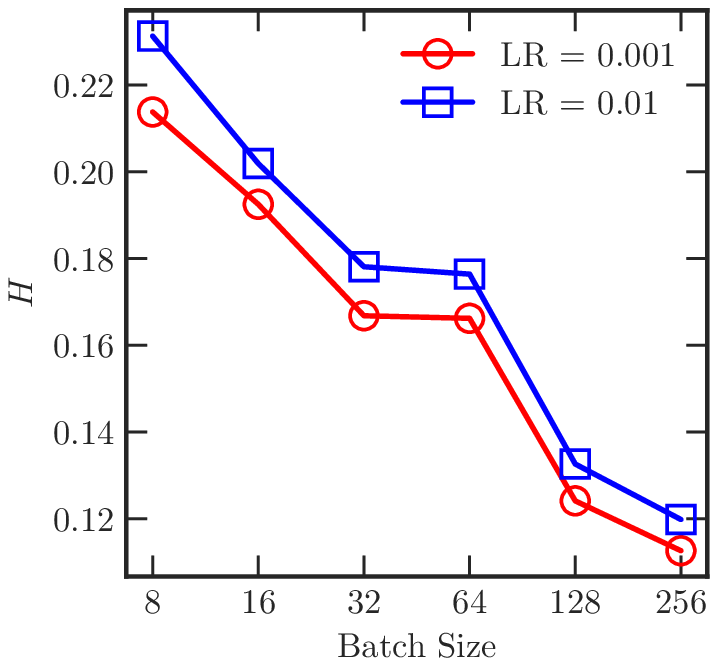}
				\caption{}
			\end{subfigure}
			\caption{Estimation of the Hurst parameter of LSTM on (a) FakeNews and (b) DisasterTweets.}
			\label{fig:lstm_fake_tweet}
		\end{figure}
		\subsection{Impacts of LSTM and Attention Mechanism}
		Unlike traditional feedforward neural networks, LSTM \cite{hochreiter1997long} has feedback connections, and excels at handling with natural
			language processing tasks. More recently, another class of models, Transformer \cite{vaswani2017attention}, which is based on attention mechanism, is proposed and is superior to LSTM in a wide spectrum of applications.
			In this section, we investigate whether or not the SGN still follows FBM in these two \emph{ad hoc} network architectures for capturing long- and short-range inter-dependence.
			We apply the LSTM on two sequential datasets, DisasterTweets\footnote{\url{https://www.kaggle.com/c/nlp-getting-started/data}} and FakeNews\footnote{\url{https://www.kaggle.com/nopdev/real-and-fake-news-dataset}},
			for the task of binary classification.
			As to the Transformer model, we use the ViT \cite{dosovitskiy2021image}, which is particularly designed to extend the Transformer model to image classification tasks.
		\par
\begin{figure}[h!]
			\centering
			\begin{subfigure}[b]{0.24\textwidth}
				\centering
				\includegraphics[width=\textwidth, clip, trim= 0 0 0 0]{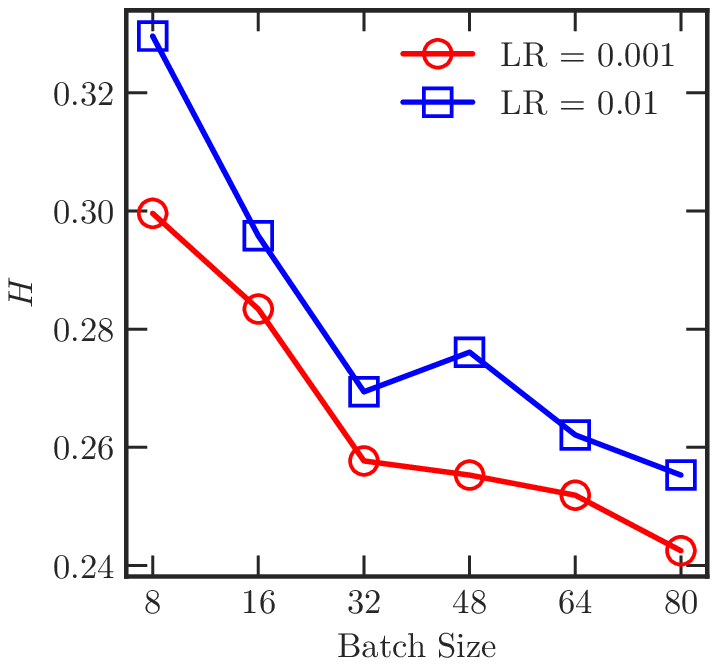}
				\caption{}
			\end{subfigure}
			\begin{subfigure}[b]{0.24\textwidth}
				\centering
				\includegraphics[width=\textwidth, clip, trim= 0 0 0 0]{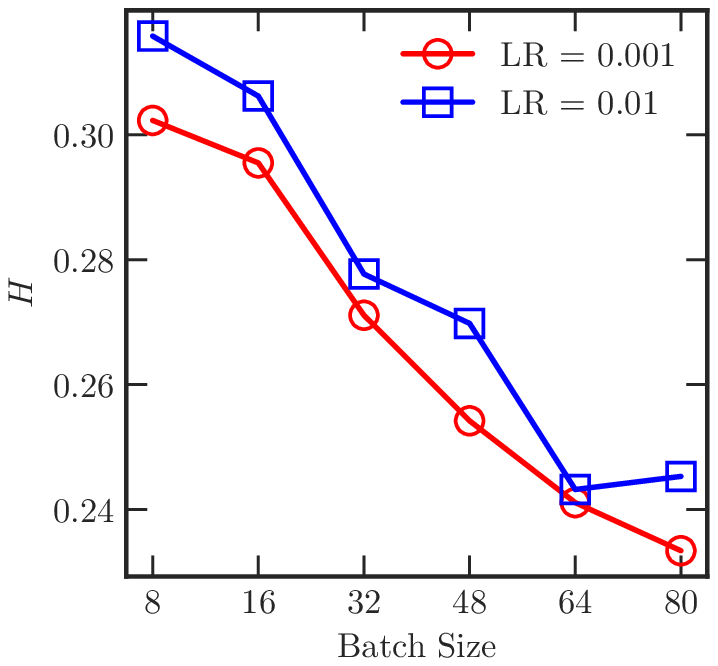}
				\caption{}
			\end{subfigure}
			\caption{Estimation of the Hurst parameter of ViT on (a) CIFAR10 and (b) CIFAR100.}
			\label{fig:vit_cifar10_cifar100}
		\end{figure}

		We still look into how the Hurst parameter changes with the batch size and learning rate.
			Due to limited GPU memory (single NVIDIA RTX2080TI), for the ViT model we conduct experiments with very small batches \{8, 16, $\cdots$, 80\} and mixed-precision training is also employed.
			As can be seen from Figs. \ref{fig:lstm_fake_tweet} and \ref{fig:vit_cifar10_cifar100}, the Hurst parameter is far away from 0.5, which suggests that SGN in LSTM and ViT is highly non-Gaussian and follows FBM as well.
		\begin{figure}[h!]
			\centering
			\begin{subfigure}[b]{0.24\textwidth}
				\centering
				\includegraphics[width=\textwidth, clip, trim= 0 0 0 0]{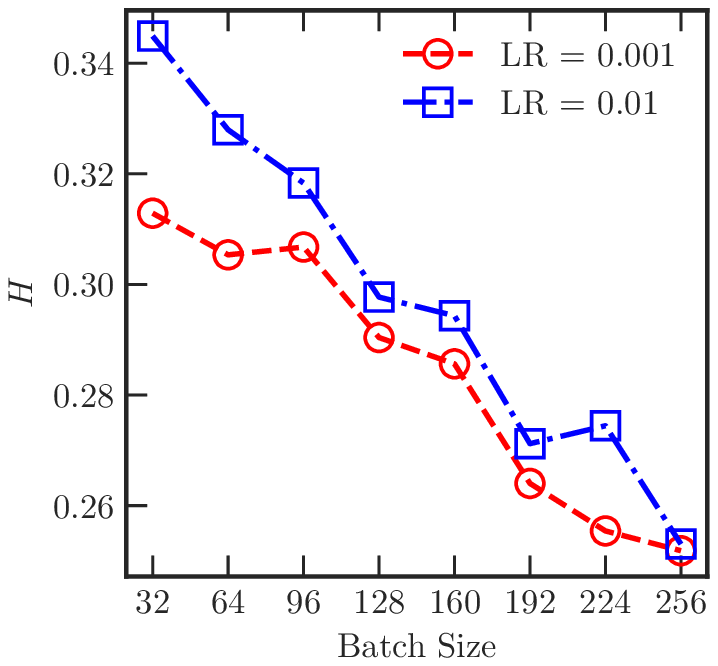}
				\caption{}
			\end{subfigure}
			\begin{subfigure}[b]{0.24\textwidth}
				\centering
				\includegraphics[width=\textwidth, clip, trim= 0 0 0 0]{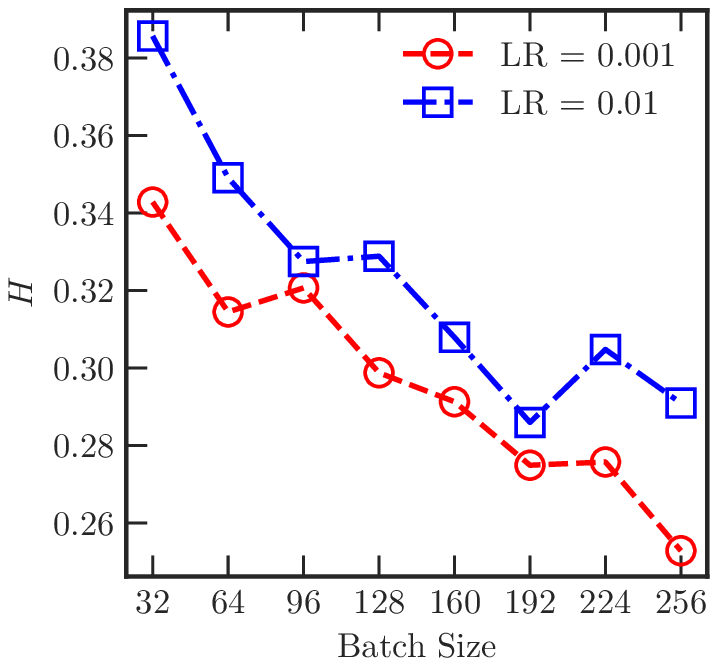}
				\caption{}
			\end{subfigure}\\
			\begin{subfigure}[b]{0.24\textwidth}
				\centering
				\includegraphics[width=\textwidth, clip, trim= 0 0 0 0]{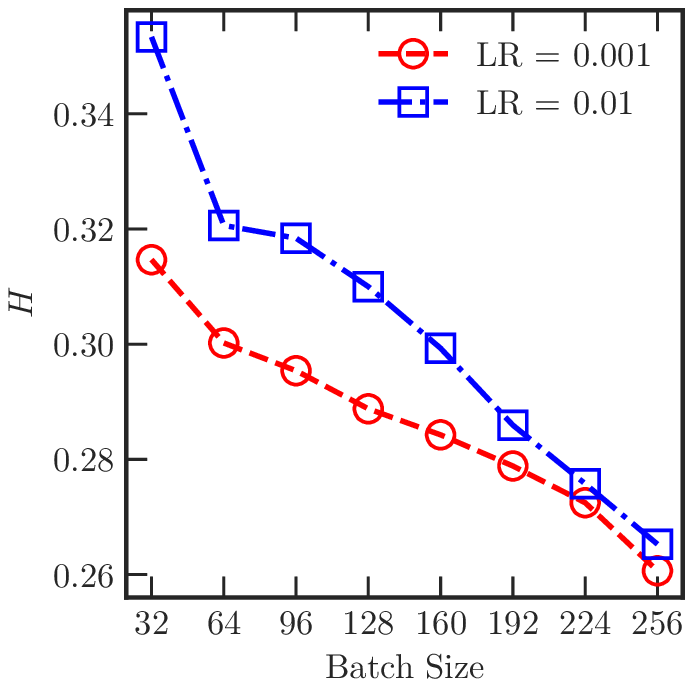}
				\caption{}
			\end{subfigure}
			\begin{subfigure}[b]{0.24\textwidth}
				\centering
				\includegraphics[width=\textwidth, clip, trim= 0 0 0 0]{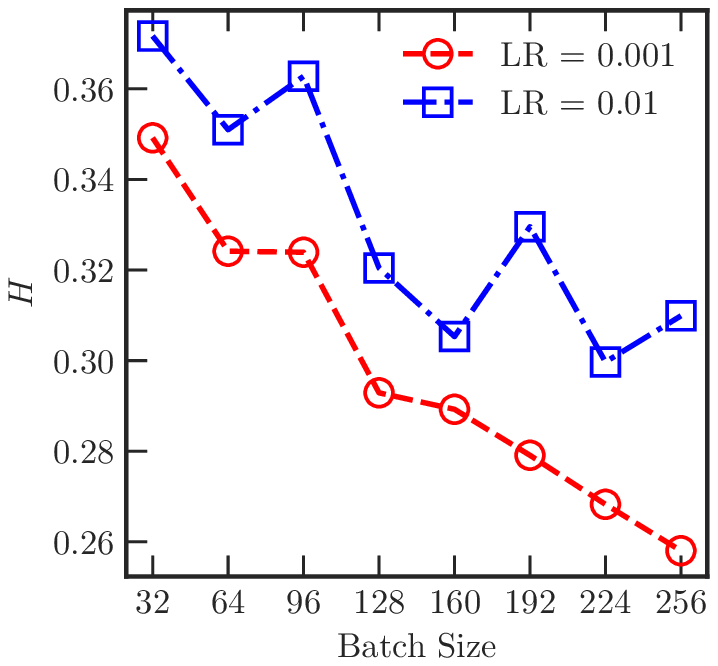}
				\caption{}
			\end{subfigure}
			\caption{Impacts of momentum and weight decaying on Hurst parameter: (a) vanilla SGD; (b) SGD + momentum; (c) SGD + weight decaying; (d) SGD + momentum + weight decaying.}
			\label{fig:momentum_weight_decay}
		\end{figure}
		\subsection{Impacts of Momentum and Weight Decaying}
		Though our theoretical analysis is based on vanilla SGD, in this section we investigate the impacts of momentum and weight decaying as both of them are indispensable ingredients to achieve better results.
			To recover the influence of different training strategies, we train ResNet18 on CIFAR10 with batches from \{32, 64, $\cdots$, 256\} and learning rates from \{0.001, 0.01\}.
			As can be seen from Fig. \ref{fig:momentum_weight_decay}, the findings of Hurst parameter on batch size and learning rate for vanilla SGD are still valid in all the settings: large learning rate and small batch size lead to large Hurst parameter.
			Moreover, training with momentum results in an increase of Hurst parameter, as shown in Fig. \ref{fig:momentum_weight_decay}(a)-(b).
			By contrast, Hurst parameter is less sensitive to weight decaying (see Fig. \ref{fig:momentum_weight_decay}(a)-(c)).
		\section{Conclusion and Future Work}
		Based on the observation that SGN is neither Gaussian nor L\'evy stable, we investigated SGD as a discretization of an FBM-driven SDE. The results of the first passage time analysis for such SDEs demonstrated that the training process prefers staying longer in wide minima. The short-range memory effect was examined and experimentally demonstrated. In the future, we intend to study the generalization properties of SGD in such a fractional noise-driven SDE framework.
		Besides, theoretical analysis on SGD with momentum should also be covered.
		\appendices
		\section{}
		\label{appendix: proof of lemma1}
		\begin{IEEEproof}[Proof of Lemma \ref{lemma1}]
			We denote by $Y_t$ the FGN corresponding to $B_t^H$, and at $t=0$, let the neural network weight $w$ be equal to $w_0$.
			Then, it follows from (\ref{eq:fou one dim}) that
			\begin{equation}
				w(t; w_0) = e^{-at} w_0 + \int_{0}^{t}  e^{-a(t-\tau)} Y_\tau \mathrm{~d}\tau.
				\label{eq:raw_solution}
			\end{equation}
			The conditional probability density function of $w$ at time $t$ is given by the Dirac delta function
			\begin{equation}
				p(w, t; w_0) = \delta\left(w - w\left(t;w_0\right)\right).
			\end{equation}
			If the distribution of $w_0$ is given, the distribution of $w$ at any time $t$ has the form
			\begin{equation}
				p(w, t) = \int  p(w_0, 0) \left<\delta\left(w - w\left(t;w_0\right)\right) \right> \mathrm{d}w_0,
				\label{eq:int_pdf}
			\end{equation}
			where $\left<\cdot\right>$ stands for $\int p_Y(\tau)\mathrm{d}\tau$, and $p_Y(\tau)$ is the probability density function of the noise $Y_t$.
			By recalling the expression
			\begin{equation}
				\delta(w) = \frac{1}{2\pi}\int  e^{iw x} \mathrm{d} x,
			\end{equation}
			it follows from (\ref{eq:raw_solution}) that
			\begin{equation}
				\left<\delta\left(w - w\left(t;w_0\right)\right)\right> = \frac{1}{2\pi} \int  \hat{G}\left(x, t\right) e^{{i(w - e^{-at}w_0)x}} \mathrm{d}x,
				\label{eq:mean_bracket}
			\end{equation}
			where 
			\begin{equation}
				\hat{G}(x, t) = \left<\exp\left(-ix\int_{0}^{t}  e^{-a(t-\tau)}Y_\tau \mathrm{d}\tau\right)\right>.
				\label{eq: average on fgn}
			\end{equation}
			Let the Fourier transform of $p(w, t)$ be $\hat{p}(x, t)$, namely, 
			\begin{equation}
				p(w, t) = \frac{1}{2\pi} \int  e^{iw x} \hat{p}(x, t) \mathrm{d}x,
			\end{equation}
			combining (\ref{eq:int_pdf}) with (\ref{eq:mean_bracket}) yield
			\begin{equation}
				\hat{p}(x, t) = \hat{G}(x, t) \hat{p}(e^{-at}x, 0).
				\label{eq:fourier_trans}
			\end{equation}
			By the stationarity of $Y_t$, we have
			\begin{equation*}
				\begin{aligned}
					<Y_1(t_1)\cdots Y_{2n+1}(t_{2n+1})> = 0,
				\end{aligned}
			\end{equation*}
			and
			\setlength{\arraycolsep}{0.0em}
			\begin{eqnarray}
				<Y_1(t_1)\cdots Y_{2n}(t_{2n})> &{}={}& \sum g_{1,2}(t_1-t_2) \nonumber\\
				&&{}\:\cdots g_{2n-1,2n}(t_{2n-1} - t_{2n}),\nonumber
			\end{eqnarray}
			where the summation operator is taken over all possible pair compositions of $1, t_1;\cdots;2n, t_{2n}$, and the total number of such compositions is $(2n-1)!!=2n! / n!2^n$.
			Moreover, the correlation function is
			\begin{equation}
				g(\tau) = H(2H-1) \tau^{2H-2}.
			\end{equation}
			Introducing the notation
			\begin{equation}
				\mathcal{Z}(t) = \int_{0}^{t} \int_{0}^{t}  e^{-at_1} e^{-at_2} g(t_1 - t_2) \mathrm{d}t_1 \mathrm{d} t_2,
			\end{equation}
			and expanding $\hat{G}(x, t)$ in a Taylor series, we obtain
			\begin{equation}
				\hat{G} (x, t) = \sum_{0}^{\infty} \frac{1}{(2n)!} \frac{(2n)!}{n!2^n}(-i)^{2n}\mathcal{Z}^n(t)x^{2n} 
			\end{equation}
			or equivalently
			\begin{equation}
				\hat{G}(x, t) = e^{-\mathcal{Z}(t)x^2}.
				\label{eq:G_hat}
			\end{equation}
			Since
			\begin{equation}
				\hat{p}(x, 0) = \int \mathrm{d}w\delta(w-w_0)e^{ixw} = e^{iw_0 x}
			\end{equation}
			and 
			\begin{equation}
				\hat{p}(e^{-at}x, 0) = \exp(-iw_0e^{-at}x),
			\end{equation}
			combining with (\ref{eq:G_hat}) yields
			\begin{equation}
				\begin{aligned}
					p(w, t) &= \frac{1}{2\pi}\int \hat{p}(x, t)e^{iw x} \mathrm{d}x\\
					&=\frac{1}{2\pi}\int \hat{G}(x, t)\hat{p}(e^{-at}x, 0)e^{iw x} \mathrm{d}x\\
					&=\frac{1}{2\pi}\int e^{-\mathcal{Z}(t)x^2}\exp(-iw_0e^{-at}x)e^{iw x} \mathrm{d}x\\
					&=\frac{1}{2\sqrt{\pi \mathcal{Z}(t)}}\exp\left[-\frac{\left(w - e^{-at}w_0\right)^2}{4\mathcal{Z}_t}\right].
				\end{aligned}
				\label{eq:analytical_exp}
			\end{equation}
			We now proceed to compute the expression $\mathcal{Z}(t)$.
			Calculating the expression $\mathcal{Z}(t)$ can be rather involved; thus, we introduce a change of variables to obtain a more tractable form
			\begin{equation}
				\begin{aligned}
					\mathcal{Z}(t) &= \int_{0}^{t} g(\tau) \int_{0}^{t-\tau} e^{-a(T+\tau)} e^{-aT} \mathrm{d}\tau \mathrm{d}T\\
					&= \sigma^2H(2H-1)\int_{0}^{t} \tau^{2H-2}e^{-a\tau}(-\frac{1}{2a})e^{-2aT}|_0^{t-\tau} \mathrm{d}\tau\\
					&= \frac{1}{2a}\sigma^2H(2H-1)\int_{0}^{t} \tau^{2H-2}\left(e^{-a\tau} - e^{a\tau}e^{-2at}\right)\mathrm{d}\tau.
				\end{aligned}
				\label{eq:Z_total}
			\end{equation}
			We note that the first term inside the brackets is
			\setlength{\arraycolsep}{0.0em}
			\begin{eqnarray}
				\int_{0}^{t} \tau^{2H-2}e^{-a\tau} \mathrm{d}\tau &{}={}& \int_{0}^{t} (\frac{\tau}{t})^{2H-2}t^{2H-2} e^{-at\frac{\tau}{t}} \mathrm{d}\tau \nonumber\\
				&{}={}&t^{2H-1} \int_{0}^{1} \tau^{2H-2}e^{-at\tau} \mathrm{d}\tau\\
				&{}={}& t^{2H-1}(\int_{0}^{\infty} \tau^{2H-2}e^{-at\tau}\mathrm{d}\tau\nonumber\\ 
				&&{-}\:\int_{1}^{\infty} \tau^{2H-2}e^{-at\tau}\mathrm{d}\tau) \nonumber\\
				&{}={}& a^{1-2H}\Gamma(2H-1) - t^{2H-1}E_{2-2H}(at), \nonumber
				\label{eq:first_term}
			\end{eqnarray}
			\setlength{\arraycolsep}{5pt}
			whereas the second term is
			\begin{equation}
				\begin{aligned}
					\int_{0}^{t} \tau^{2H-2}e^{a\tau}e^{-2at} \mathrm{d}\tau &= e^{-2at}\int_{0}^{t} \tau^{2H-2}e^{a\tau} \mathrm{d}\tau\\
					&= e^{-2at} t^{2H-1} \int_{0}^{1} \tau^{2H-2}e^{at\tau} \mathrm{d}\tau\\
					&=\frac{t^{2H-1}}{2H-1}e^{-2at}M(2H-1, 2H, at).
				\end{aligned}
				\label{eq:second_term}
			\end{equation}
			Combining (\ref{eq:analytical_exp})--(\ref{eq:second_term}) concludes the proof.
		\end{IEEEproof}
		\section{}
		\label{appendix: proof of theorem1}
		\begin{IEEEproof}[Proof of Theorem \ref{theorem2}]
			On one hand, when $t$ is sufficiently large,
			\begin{equation}
				\int_{0}^{t} \tau^{2H-2}e^{-a\tau} \mathrm{d}\tau \approx a^{1-2H} \Gamma(2H-1).
				\label{eq: appendix b first term}
			\end{equation}
			On the other hand, since
			\begin{equation}
				M(2H-1, 2H, at) \approx \frac{2H-1}{at}e^{at} + \Gamma(2H)(-at)^{1-2H},
			\end{equation}
			we have
			\begin{equation}
				\int_{0}^{t} \tau^{2H-2}e^{a\tau}e^{-2at} \mathrm{d}\tau \approx \frac{t^{2H-2}}{a} e^{-at}.
				\label{eq: appdix b second term}
			\end{equation}
			Combining (\ref{eq: appdix b second term}), (\ref{eq: appendix b first term}) and (\ref{eq:Z_total}), it is direct to conclude that
			\setlength{\arraycolsep}{0.0em}
			\begin{eqnarray}
				\mathcal{Z}(t) &{}\approx{}& \frac{H(2H-1)} {2a^{2H}} \Gamma(2H-1) \sigma^2 \\
				&&{-}\:\frac{\sigma^2}{2a^2}H(2H-1) t^{2H-1} \frac{e^{-at}}{at} \nonumber\\
				&{}={}&\frac{H(2H-1)} {2a^{2H}} \Gamma(2H-1) \sigma^2 \left(1- \frac{t^{2H-2}e^{-at}}{\Gamma(2H-1)a^{2-2H}}\right). \nonumber
			\end{eqnarray}
			Using the Taylor expansion of exponential function, we have
			\setlength{\arraycolsep}{0.0em}
			\begin{eqnarray}
				p(w, t) &{}={}& \frac{1}{2\sqrt{\pi\mathcal{Z}_t}}\exp\left[-\frac{\left(w - e^{-at}w_0\right)^2}{4\mathcal{Z}_t}\right] \\
				&{}\approx{}& \frac{a^H}{\sqrt{2\pi \Gamma(2H-1)H(2H-1)\sigma^2}}\nonumber\\
				&&{\times}\:\left(1-\frac{a^{2H-2}t^{2H-2}e^{-at}}{2\Gamma(2H-1)}+\cdots\right)\nonumber\\
				&&{\times}\:\left(1-\frac{\left(w - e^{-at}w_0\right)^2}{4\mathcal{Z}_t} + \cdots \right)\\
				&{}\approx{}& C_1+ C_2 \frac{a^{3H}t^{2H-2}e^{-at}}{\sigma}(1+C_3w+C_4w^2+\cdots).\nonumber
			\end{eqnarray}
			where $C_1, C_2, C_3, C_4$ are constants.
			Recall that the survival probability $s(t)$ is the probability that the random walker remains in the domain up to time $t$, and it can be expressed as
			\begin{equation}
				s(t) = \int  p(w, t) \mathrm{d}w.
			\end{equation}
			Hence, the first passage time density is directly obtained from
			\begin{equation}
				p(t)=-\frac{\mathrm{d} s(t)}{\mathrm{d}t} \approx \mathcal{O}(\frac{a^{3H}t^{2H-2}e^{-at}}{\sigma}).
				\label{eq:first passage time dnesity}
			\end{equation}
		\end{IEEEproof}
		\bibliography{sgd_2020}
		\bibliographystyle{IEEEtran}
		\begin{IEEEbiography}[{\includegraphics[width=1.25in,height=1.25in,clip,keepaspectratio]{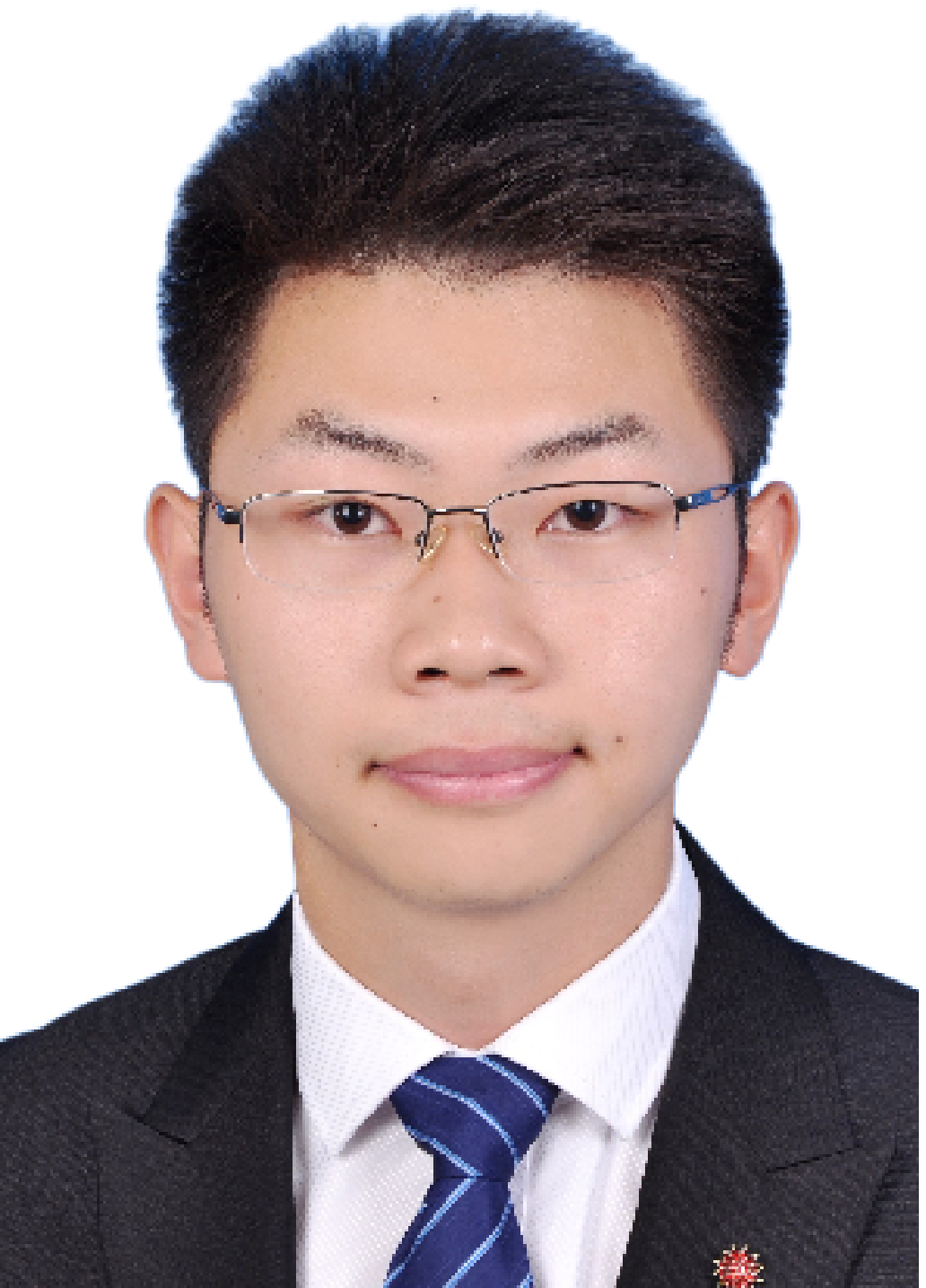}}]{Chengli Tan}
			received the B.S. degree in information and computing science and the M.S. degree in statistics from Xi’an Jiaotong University, Xi’an, China, in 2014 and 2017, where he is now pursuing the Ph.D. degree. His current research interests include adversarial learning, Bayesian nonparametrics, and stochastic optimization.
		\end{IEEEbiography}
		
		\begin{IEEEbiography}[{\includegraphics[width=1.25in,height=1.5in,clip,keepaspectratio]{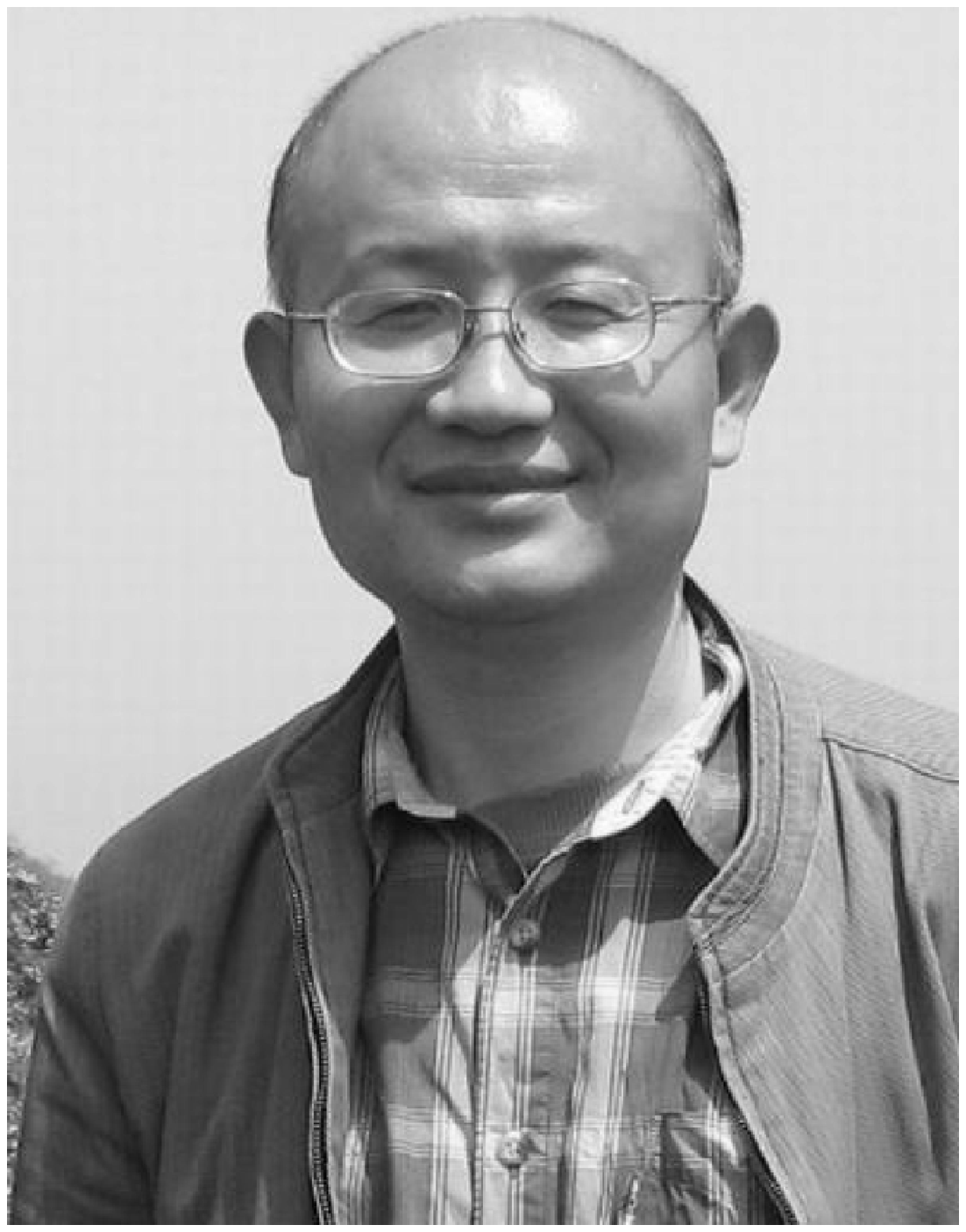}}]{Jiangshe Zhang}%
			received the B.S., M.S., and Ph.D. degrees in computational mathematics from Xi’an Jiaotong University, Xi’an, China, in 1984, 1987, and 1993, respectively. He is currently the Director of the Institute of Machine Learning and Statistical Decision Making, Xi’an Jiaotong University, where he is a Professor	with the Department of Statistics. He is also the Vice-President of the Xi’an International Academy for Mathematics and Mathematical Technology, Xi’an, China. He has authored and co-authored one monograph and over 100 journal papers. His research interests include statistical computing, deep learning, cognitive representation, and statistical decision-making.
			\par
			Prof. Zhang received the National Natural Science Award of China (Third Place) and the First Prize in Natural Science from the Ministry of Education of China in 2007. He served as the President of the Shaanxi Mathematical Society and the Executive Director of the China Mathematical Society.
		\end{IEEEbiography}
		
		\begin{IEEEbiography}[{\includegraphics[width=1.25in,height=1.25in,clip,keepaspectratio]{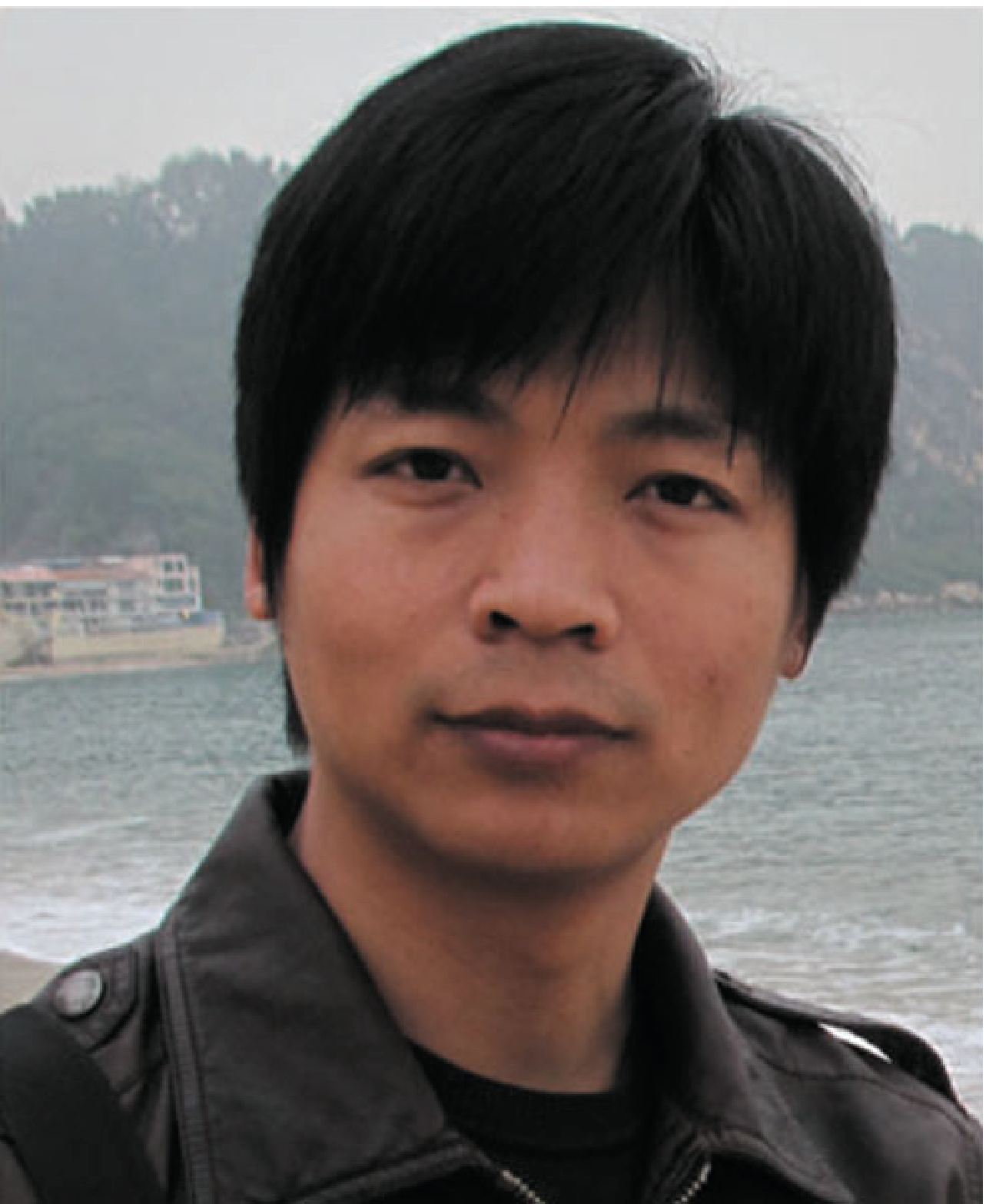}}]{Junmin Liu}%
			received a Ph.D. in applied mathematics from Xi'an Jiaotong University, Xi'an, China in 2013. He is currently an Associate Professor in the School of Mathematics and Statistics, Xi'an Jiaotong University, Xi'an, China. From 2014 to 2015, he was a visiting scholar at the University of Maryland, College Park. His research interests are focused on machine learning and computer vision.
		\end{IEEEbiography}
		\vfill
	\end{document}